\documentclass{article}

    \PassOptionsToPackage{numbers, compress}{natbib}

    \usepackage[preprint]{neurips_2022}

\usepackage[utf8]{inputenc} 
\usepackage[T1]{fontenc}    
\usepackage{hyperref}       
\usepackage{url}            
\usepackage{booktabs}       
\usepackage{amsfonts}       
\usepackage{nicefrac}       
\usepackage{microtype}      
\usepackage{xcolor}         

\usepackage{multirow}
\usepackage{graphicx}
\usepackage{wrapfig}
\usepackage{subcaption}
\usepackage{makecell}
\usepackage{bbding}
\usepackage{amsmath}
\usepackage{colortbl}
\usepackage{xr}
\makeatletter
\newcommand*{\addFileDependency}[1]{
  \typeout{(#1)}
  \@addtofilelist{#1}
  \IfFileExists{#1}{}{\typeout{No file #1.}}
}
\makeatother

\newcommand*{\myexternaldocument}[1]{
    \externaldocument{#1}
    \addFileDependency{#1.tex}
    \addFileDependency{#1.aux}
}
\myexternaldocument{appendix}

\title{Effective Vision Transformer Training: A Data-Centric Perspective}

\author{%
  Benjia Zhou$^{1,2}$\thanks{Work done during an internship at Alibaba Group.}, 
  Pichao Wang$^{2\dagger}$, 
  Jun Wan$^{1,3,4}$\thanks{Corresponding author.}, Yanyan Liang$^{1}$, Fan Wang$^{2}$ \\
  $^{1}$Macau University of Science and Technology \quad $^{2}$Alibaba Group \\
 $^{3}$NLPR, Institute of Automation, Chinese Academy of Sciences, Beijing, China \\
 $^{4}$School of Artificial Intelligence, University of Chinese Academy of Sciences, Beijing, China 
}

\begin{document}
\def\etal{\emph{et al}.}
\def\eg{\emph{e}.\emph{g}.}
\def\ie{\emph{i}.\emph{e}.}

\maketitle

\begin{abstract}
Vision Transformers (ViTs) have shown promising performance compared with Convolutional Neural Networks (CNNs), but the training of ViTs is much harder than CNNs. Current ViT training suffers from sub-optimal due to the complexity of learning curves. In this paper, we define several metrics, including Dynamic Data Proportion (DDP) and Knowledge Assimilation Rate (KAR), to investigate the training process, and divide it into three periods accordingly: \textbf{formation}, \textbf{growth} and \textbf{exploration}. 
As revealed from our observations, the same set of training examples being fed into the different training stages leads to insufficient learning, since the number of ``effective'' training examples for networks differs widely across these stages.
In particular, at the last stage of training, we observe that only a tiny portion of training examples is used to optimize the model. Given the data-hungry nature of ViTs, we thus ask a simple but important question: is it possible to provide abundant ``effective'' training examples at EVERY stage of training? To address this issue, we need to address two critical questions, \ie, how to measure the ``effectiveness'' of individual training examples, and how to systematically generate enough number of ``effective'' examples when they are running out. To answer the first question, we find that the ``difficulty'' of training samples can be adopted as an indicator to measure the ``effectiveness'' of training samples. To cope with the second question, we propose to dynamically adjust the ``difficulty'' distribution of the training data in these evolution stages. To achieve these two purposes, we propose a novel data-centric ViT training framework to dynamically measure the ``difficulty'' of training samples and generate ``effective'' samples for models at different training stages. Furthermore, to further enlarge the number of ``effective'' samples and alleviate the overfitting problem in the late training stage of ViTs, we propose a patch-level erasing strategy dubbed \textbf{PatchErasing}. Extensive experiments demonstrate the effectiveness of the proposed data-centric ViT training framework and techniques. 
\end{abstract}

\section{Introduction}
In the last decade, Convolutional Neural Networks (CNNs) driven by large-scale data have revolutionized the field of computer vision. Recently, Vision Transformers (ViTs), first introduced by Dosovitskiy \etal~\cite{dosovitskiy2020image}, have progressively emerged as the alternative architectures in computer vision. However, training of ViTs is much harder than CNN as the former tends to be more data-hungry ~\cite{touvron2021training,wang2021scaled} due to the lack of inductive bias, 
making ViTs vulnerable to over-fitting than CNNs. 
Although various architectures~\cite{han2021transformer,d2021convit,yuan2021incorporating,xu2021co,chen2021crossvit,graham2021levit,li2021localvit,wang2021kvt,guo2021cmt,fan2021multiscale,rao2021dynamicvit,el2021xcit,chen2021psvit,huang2021shuffle,mobilevit,cswin,hrformer,zhou2021elsa} have been continuously suggested to improve the ViTs, these models still suffer from suboptimal performance under the limited training data. 
Therefore, follow-up efforts start to pay attention to data-centric solutions, which is a promising direction for training ViTs. 
As a representative, data augmentation~\cite{shorten2019survey,touvron2021training,balestriero2022effects,xu2022comprehensive}, a data-driven and informed regularization strategy that artificially increases the number of training samples to improve the generalization ability of models, has been widely explored for better training of ViTs.

\begin{wrapfigure}[17]{r}{0.55\textwidth}
    \centering
    \includegraphics[width=1.0\linewidth]{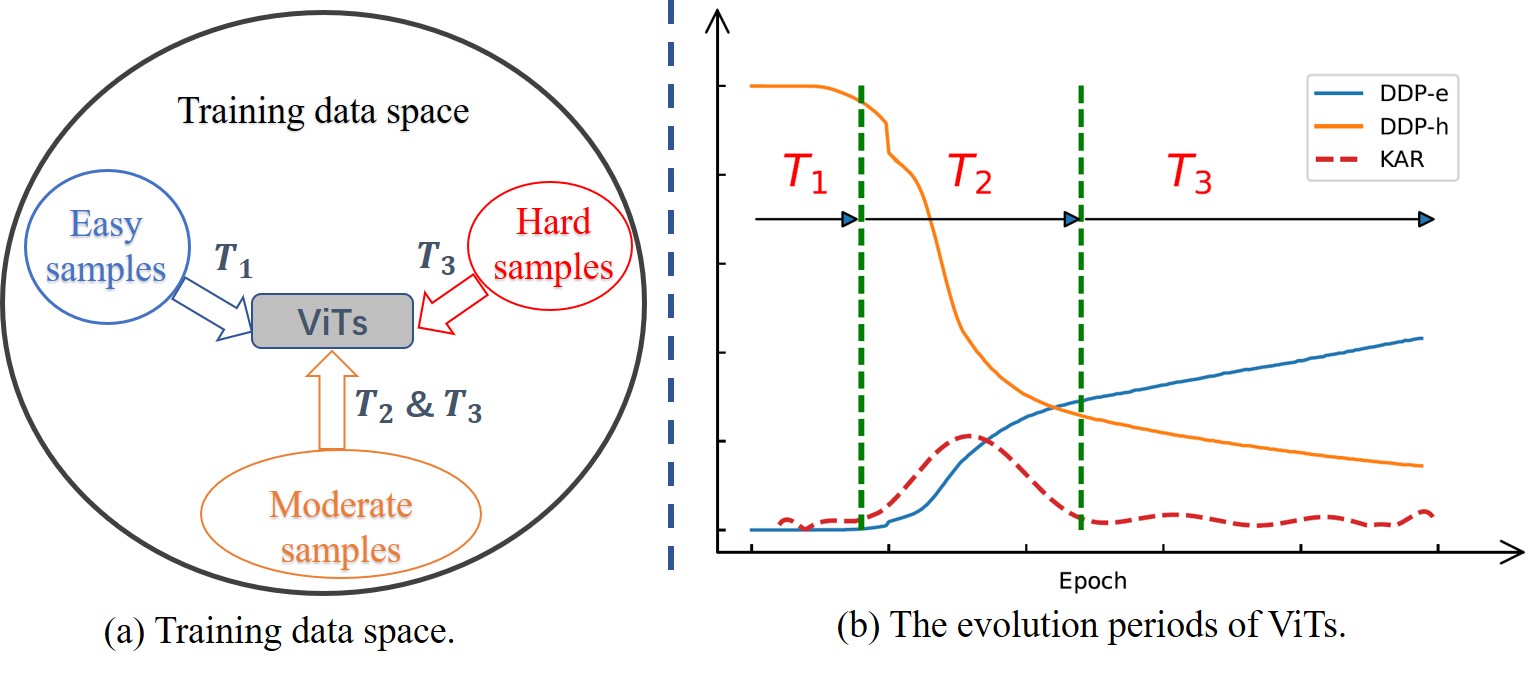}
    \caption{\textbf{The learning process of ViTs.} Wherein $\mathrm{DDP}$-e and $\mathrm{DDP}$-h represent Dynamic Data Proportion of easy and hard samples, respectively; KAR represents Knowledge Assimilation Rate. According to our investigation, ViTs training can be divided into three distinct periods $T_1$, $T_2$ and $T_3$.}
    \label{fig:learning}
\end{wrapfigure}
However, traditional ViTs training methods exhibit an inflexible training paradigm, which is data-unfriendly. Based on our observations, the training process can be divided into several distinct stages. To investigate this problem, we define several metrics, including Dynamic Data Proportion (DDP) and Knowledge Assimilation Rate (KAR), and divide the training process into three stages accordingly: formation, growth and exploration, as shown in Figure~\ref{fig:learning} (a,b), following an easy sample to hard sample learning process. The formation stage ($T_1$) is the period when the model forms as a fledgling model driven by easy samples; the growth stage ($T_2$) is when the model absorbs knowledge from moderate samples; and the exploration stage($T_3$) is when the model strives to explore more new knowledge from moderate and hard samples.

We find that even though the same set of training examples is fed into the training process at different stages, the number of ``effective" training examples is very scarce. For example, at the last stage of training, we observe that only hard and moderate training examples are used to optimize the model, leaving the training loss of easy samples nearly unchanged. Given the data-hungry nature of ViTs, we raise an interesting question: how to provide as many ``effective'' training examples at every stage of training as possible? To address this issue, we need to solve two other questions: how to measure the ``effectiveness" of individual training examples, and how to generate enough number of ``effective'' examples upon need.

For the first question, it is found that the ``difficulty'' of training samples can be adopted to measure the ``effectiveness'' of training samples.  
To generate ``effective'' samples, we propose to dynamically adjust the ``difficulty'' of training samples through a dynamic MixUp technique at different training stages. More concretely, we propose a novel data-centric ViT training framework, which contains a data samples processing mechanism that can solve many data-centric tasks, herein the measurement of sample ``difficulty" and generation of ``effective'' examples. The model trained from prior iteration can provide extensive data operation guidance for the current model to learn.
Different from curriculum learning~\cite{bengio2009curriculum,graves2017automated}, we divide the evolution period of training into three stages explicitly, and flexibly manipulate the ``difficulty'' distribution of training data according to the distinguishing properties of these stages.
To further enlarge the ``effective'' samples and alleviate the overfitting problem in the late training stage of ViT, a patch-level erasing method namely PatchErasing is customized for ViTs. It relieves the ViTs from overfitting to some local patch tokens in the late $T_3$ period by randomly erasing some patches in the image and regularizing the corresponding label vector. And we experimentally demonstrate that it can improve the generalization ability of ViTs.

In summary, we make the following contributions in this work: (i) We thoroughly analyze the properties of ViT training, and by defining two new metrics, DDP and KAR, we divide the training into three periods, unveiling the process of training from easy to hard samples.
(ii) We propose to measure the ``difficulty" of training samples as the indicator of ``effectiveness" during training, and introduce a novel data-centric ViT training framework to dynamically measure the ``effectiveness" of training samples and generate `` effective" samples for different training stages. (iii) A new patch-level erasing strategy, PatchErasing, is invented to further enlarge the number of ``effective'' samples and relieve the overfitting problem in the late training stage.

\begin{figure}[ht]
  \centering
    \begin{subfigure}{0.30\textwidth}
    \includegraphics[width=\linewidth]{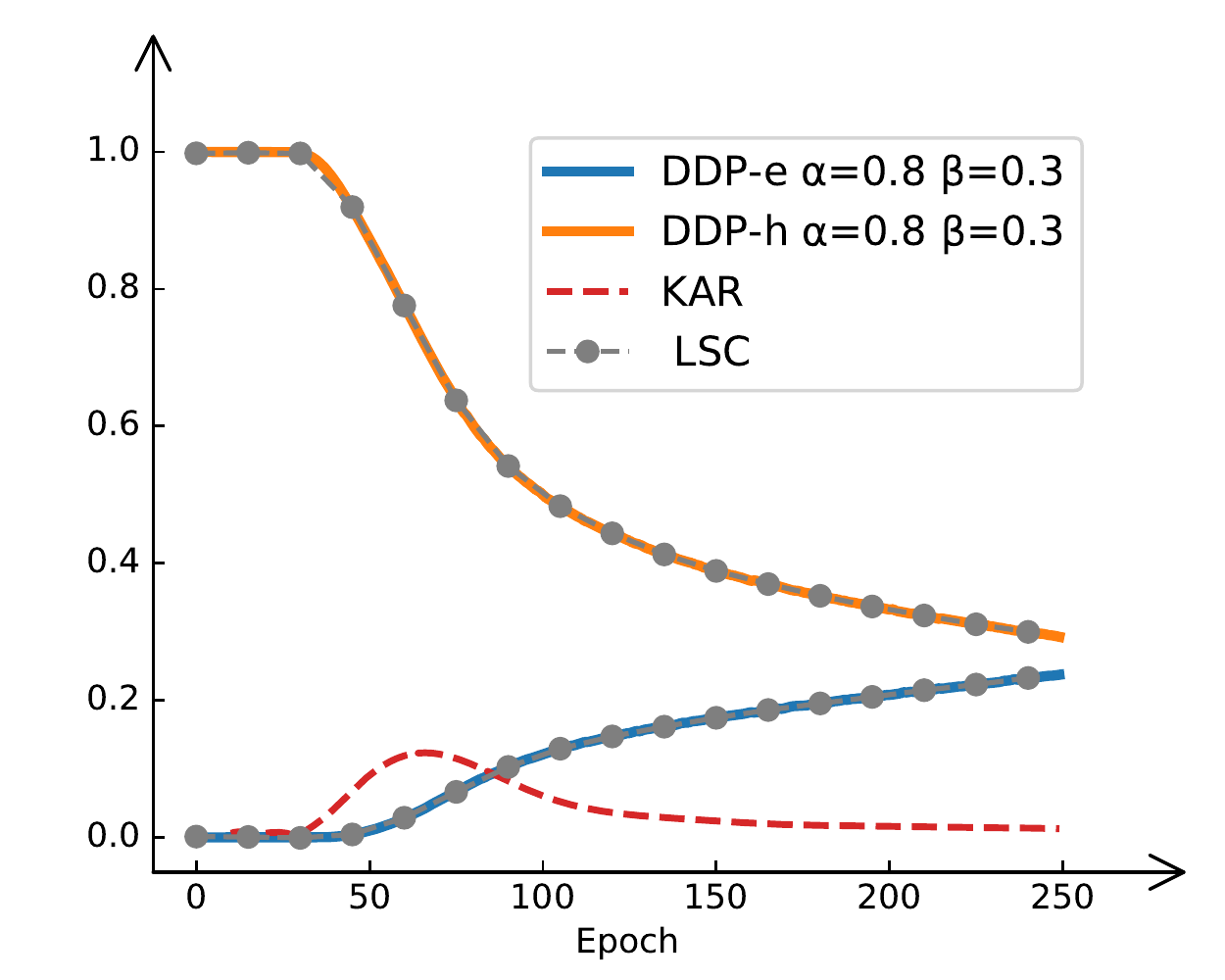}
    \caption{DeiT-T.}
    \label{subfig:deit_tiny}
    \end{subfigure}
    \hfill
    \begin{subfigure}{0.30\textwidth}
    \includegraphics[width=\linewidth]{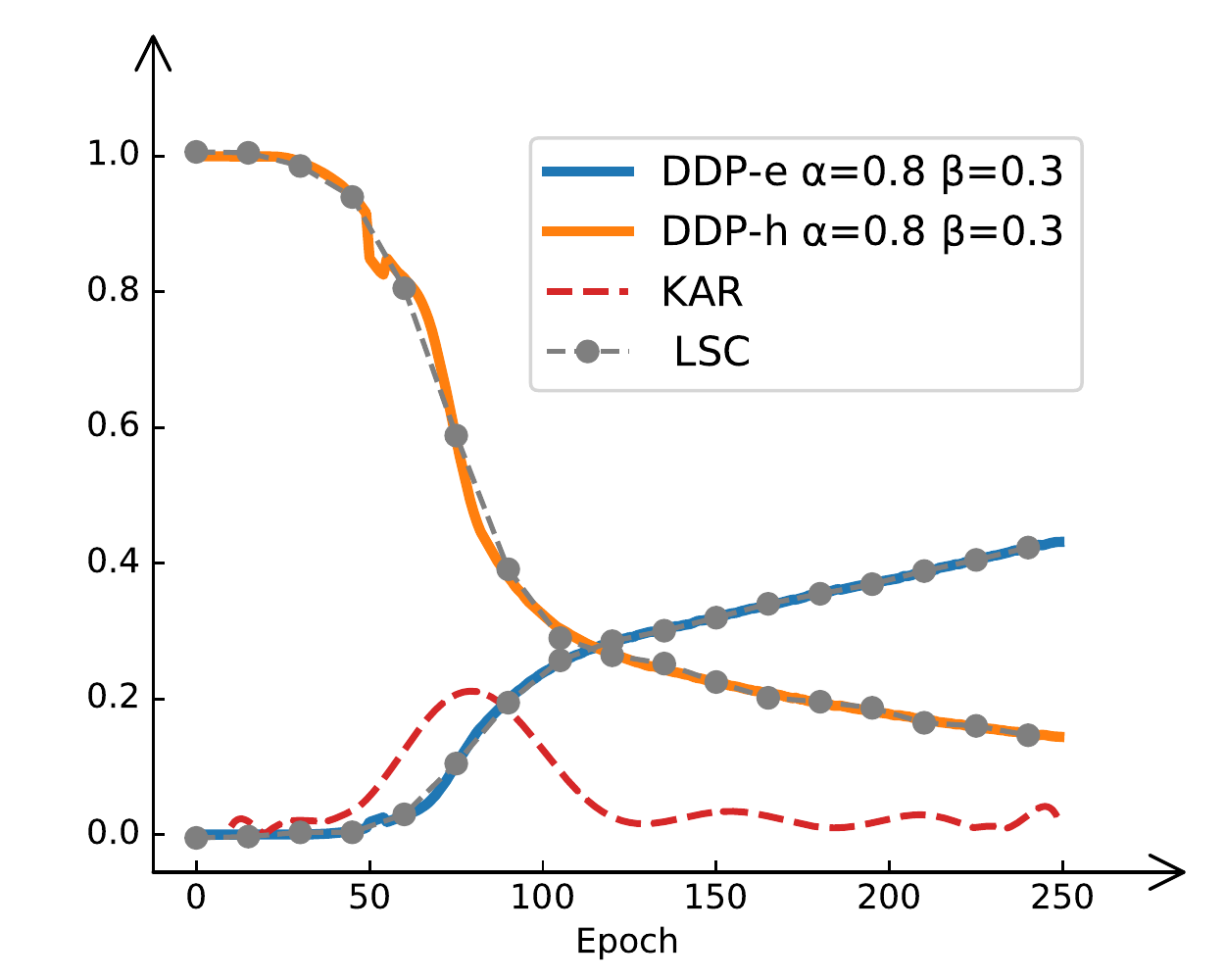}
    \caption{DeiT-S.}
    \label{subfig:deit_small}
    \end{subfigure}
    \hfill
    \begin{subfigure}{0.30\textwidth}
    \includegraphics[width=\linewidth]{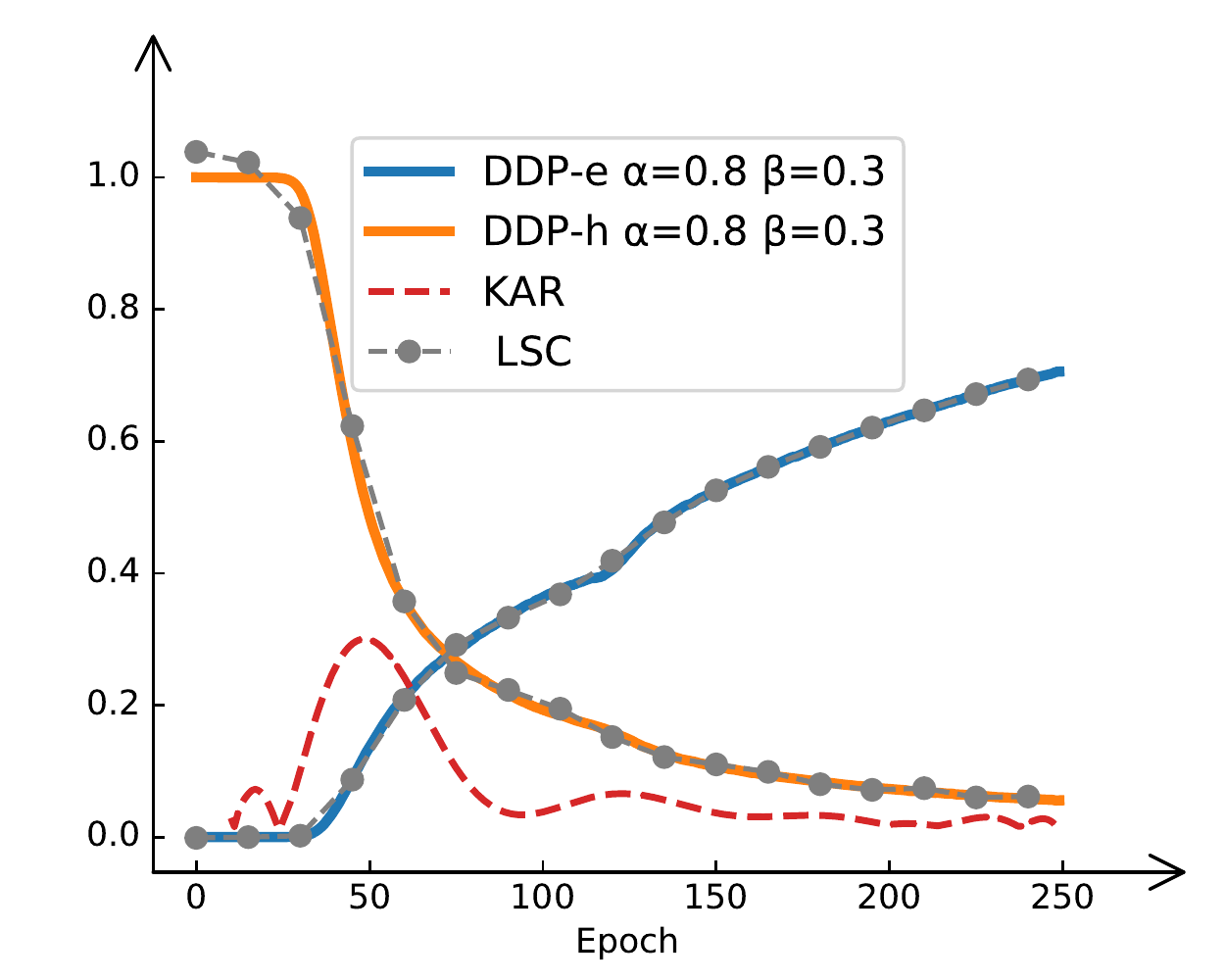}
    \caption{DeiT-B.}
    \label{subfig:deit_base}
    \end{subfigure}

  \caption{\textbf{The evolution period of DeiT.} Wherein the LSC represents the Least-Squares fitting Curve, which is mainly used to calculate the Knowledge Assimilation Rate (KAR) of the network. 
  }
  \label{fig:deitperiod}
\end{figure}

\section{Related Work}
Generally speaking, there are two directions to improve the ViT training, namely, model-centric and data-centric. We will briefly survey works towards these two directions. Please refer to \cite{khan2021transformers,xu2022comprehensive} for a comprehensive survey of ViTs and data augmentation.

\subsection{Model-Centirc ViTs Training}
Training ViTs is much harder than training CNNs, largely due to the lack of inductive bias. To deal with this problem, several works improve the ViTs training from the view of model design. CvT~\cite{wu2021cvt}, CoaT~\cite{xu2021co} and Visformer~\cite{Zhengsu21} introduce convolution operations, safely removing the position embedding; DeepViT~\cite{zhou2021deepvit} and CaiT~\cite{touvron2021going} investigate the unstable training problem, and propose the re-attention and re-scale techniques for stable training; to accelerate the convergence of training, ConViT~\cite{d2021convit}, PiT~\cite{heo2021rethinking}, CeiT~\cite{yuan2021incorporating}, and LocalViT~\cite{li2021localvit} introduce convolutional bias to speedup the training; \textit{conv-stem} is adopted in LeViT~\cite{graham2021levit}, EarlyConv~\cite{xiao2021early}, CMT~\cite{guo2021cmt}, VOLO~\cite{yuan2021volo} and ScaledReLU~\cite{wang2021scaled} to improve the robustness of training ViTs. All of these methods improve ViT training by designing new models, but our work investigates ViTs training from the view of training data.

\subsection{Data-Centirc ViTs Training}
Different from model-centric ViTs training, several works are proposed to improve the performance from the view of data.  For example, DeiT~\cite{touvron2021training} adopts several data augmentations~\cite{cubuk2018autoaugment,cubuk2020randaugment,zhong2020random} and training techniques to extend ViT~\cite{dosovitskiy2020image} to a data-efficient version; PatchViT~\cite{gong2021improve} proposes several loss functions to promote patch diversification for training wider and deeper ViTs; LV-ViT~\cite{jiang2021token} adopts several techniques including MixToken and Token Labeling for better training and feature generation. Recently, methods based on masked image modeling, such as SimMIM~\cite{xie2021simmim} and MAE~\cite{He_2022_CVPR}, are proposed to improve the self-supervised ViTs training by masking a large ratio of patches. All of these methods adopt a one-stage training pipeline, without deep analysis of the training process. In this paper, we divide the training process into three periods and design a novel unified data-centric ViTs training framework to dynamically measure the ``effectiveness" of training samples and generate as many ``effective'' samples as possible according to training stages.

\section{Data-Centric Observations} 
According to Curriculum Learning (CL)~\cite{bengio2009curriculum}, deep networks follow a learning process from easy samples to hard samples. In this paper, we insight into this phenomenon under the mainstream ViT-based architectures on ImageNet1K~\cite{deng2009imagenet}, and specifically divide the training process into several distinct stages. To this end, we first define several metrics, including Dynamic Data Proportion (DDP) and Knowledge Assimilation Rate (KAR).
\paragraph{DDP}
The Dynamic Data Proportion (DDP) reflects the proportion of easy and hard samples in the training data space.
Suppose $\mathcal{D}$ denotes training data space, $f(\cdot, \Theta_t)$ means the model parameterized by $\Theta$ during the $t$-th training schedule. So the classification probabilities predicted by $f$ can be denoted as $\mathrm{p}_t=[p_{t, 1}, p_{t, 2}, \cdots, p_{t, C}]$. Each element in $\mathrm{p}_t$ can be obtained by the softmax function:
\begin{equation}
    p_{t, i} = \frac{\mathrm{exp(z_{t, i})}}{\sum_{j=1}^{C}{\mathrm{exp}(z_{t, j})}}, \quad
    z_{t} = f(\mathrm{x}, \Theta_t)
\end{equation}
where $z_t$ represents the logits reasoned by $f$ from data examples $\mathrm{x}$, and $C$ is the number of classes. 
For a probability distribution $\mathrm{p}_t$, $p_{t, k} = \mathrm{max}(\mathrm{p}_t)$ is the maximum prediction value in it. According to the $p_{t, k}$, we can partition the data space $\mathcal{D}$ into three subsets of easy samples $\mathcal{D}_{t, e}$ that $p_{t, k} > \alpha$, moderate samples $\mathcal{D}_{t, m}$ that $\alpha \geq p_{t, k} \geq \beta$, and hard samples $\mathcal{D}_{t, h}$ that $p_{t, k} < \beta$ during training. $\alpha$ and $\beta$ denote the thresholds for separating hard and easy samples (generally we take $\alpha$=0.8, $\beta$=0.3). As a result, the static training set $\mathcal{D}$ can be reformulated as a controllable dynamic training set $\{\mathcal{D}_{t, e}, \mathcal{D}_{t, m}, \mathcal{D}_{t, h}\}$.
After that, we can formalize the $\mathrm{DDP}$ of easy and hard samples at training time $t$ as: 
\begin{equation}
        \mathrm{DDP}_{t,e} = \frac{\vert\mathcal{D}_{t, e}\vert}{\vert\mathcal{D}_N\vert}, \quad
        \mathrm{DDP}_{t,h} = \frac{\vert\mathcal{D}_{t, h}\vert}{\vert\mathcal{D}_N\vert}
\end{equation}
where $\vert\cdot\vert$ is the cardinality and $N$ is the number of examples in the mini-batch. Here we only count the proportion of easy and hard samples, because they are on separate sides of the data space, which can more intuitively reflect the state of the model.

\paragraph{KAR}
The Knowledge Assimilation Rate (KAR) reflects the rate and ability of the models to assimilate knowledge from $\mathcal{D}$. Intuitively, it reflects the sum of the change rates of $\mathrm{DDP}_{e}$ and $\mathrm{DDP}_{h}$ at each moment.
For convenience, we average the partial derivatives of $\mathrm{DDP}_{t,e}$ and $\mathrm{DDP}_{t,h}$ at time $t$ as the Knowledge Assimilation Rate (KAR) of model $f$, which is formulated as
\begin{equation}
    \mathrm{KAR}=\frac{1}{2}(\vert \frac{\partial~\mathrm{DDP}_{t,e}}{\partial t} \vert +  \vert \frac{\partial~\mathrm{DDP}_{t,h}}{\partial t}\vert).
\end{equation}
To calculate the KAR values of the model at each epoch, as shown in Figure.~\ref{fig:deitperiod}, we leverage the Least-Squares to fit the functions of $\mathrm{DDP}_{t,e}$ and $\mathrm{DDP}_{t,h}$, in which $t \in \{1, 2, \cdots, epoch\}$. 

According to the DDP and KAR, the evolution of the network can be roughly divided into three stages accordingly: formation ($T_1$: the period before the KAR increases), growth ($T_2$: the period when the KAR increases and then declines), and exploration ($T_3$: the period after the KAR declines), as shown in Figure~\ref{fig:learning} (b). And the performance of models with different capacities differs widely in these three evolutionary periods.
\begin{table}
  \caption{\textbf{Exploration of effective training examples in different evolution periods.}  In this table, we further investigate the effect of data settings in different difficulties on the three stages of network training. Top-1 accuracy is reported in this table. 
  ME means making samples easy, while MH means making samples hard. In practice, we can achieve ME and MH by adjusting the interpolation strength in MixUp.
 }
  \label{tab:evolu_invest}
  \centering
  \begin{tabular}{ccccc|cccccc}
    \toprule
     ME & MH & \makecell{$T_1$}  & \makecell{$T_2$}  & \makecell{$T_3$}   &  DeiT-T & DeiT-S & DeiT-B &  Swin-T & Swin-S & Swin-B\\
    
    \midrule
    - & - &  &  &  & 72.2 & 79.8 & 81.8 & 81.3 & 83.0 & 83.5\\
    \CheckmarkBold &  &  \CheckmarkBold &    &    & 72.5 & 79.9 & 81.8  &   81.4 & 83.0 & 83.5\\
    \CheckmarkBold &  & \CheckmarkBold & \CheckmarkBold &  & 72.7 & 80.0 &  81.6  & 81.5 & 82.8 & 83.3  \\
    \CheckmarkBold &  & \CheckmarkBold & \CheckmarkBold & \CheckmarkBold &  \textbf{72.9}   &  79.4 &  81.3  &  81.1 & 82.4 & 82.8 \\
    \midrule
    & \CheckmarkBold &  &  & \CheckmarkBold & 71.6  & \textbf{80.2} &  82.1 & 81.4 & 83.3 & 83.7\\
    & \CheckmarkBold &  & \CheckmarkBold & \CheckmarkBold &  71.0 & 80.0 &  \textbf{82.2} &  \textbf{81.6} & \textbf{83.4} & \textbf{83.8}\\
    & \CheckmarkBold & \CheckmarkBold & \CheckmarkBold & \CheckmarkBold &  69.5  & 79.7  &  81.7 &  81.0 & 82.9 & 83.6\\
    \bottomrule
  \end{tabular}
\end{table}

\paragraph{Formation Period} This is a relatively short period of time, the models at this period are gradually full-fledged by rapidly assimilating the knowledge in easy examples $\mathcal{D}_{t, e}$. 
Interestingly, we observe that the time spent in this period is basically the same for networks of different capacities as shown in Figure \ref{fig:deitperiod}, which means that the ViTs with different parameter scales have a similar initial learning ability to assimilate knowledge from easy examples. Of course, due to their different capacities, small-capacity networks, such as DeiT-tiny, only enjoy a smaller easy data space $\mathcal{D}_{t, e}$. 

As can be seen in Table~\ref{tab:evolu_invest}, taking the DeiT as an example, making samples easy in this period can bring some gains for DeiT-T and DeiT-S, while not working for DeiT-B. For another, making samples hard during this period results in a consistent performance degeneration both on DeiT-T, DeiT-S, and DeiT-B. It is in line with the learning process of CL~\cite{xiang2020learning} from easy examples to hard examples.
Nevertheless, we find that data manipulations during the $T_1$ period only caused slight performance fluctuations, which reflects the fact that the models in this period are relatively fragile, resulting in limited manipulative space on training data. Therefore, in this period, we attempt to keep the ``difficulty'' distribution of training examples at a relatively low level.

\begin{wrapfigure}[19]{r}{0.55\textwidth}
    \centering
    \includegraphics[width=1.0\linewidth]{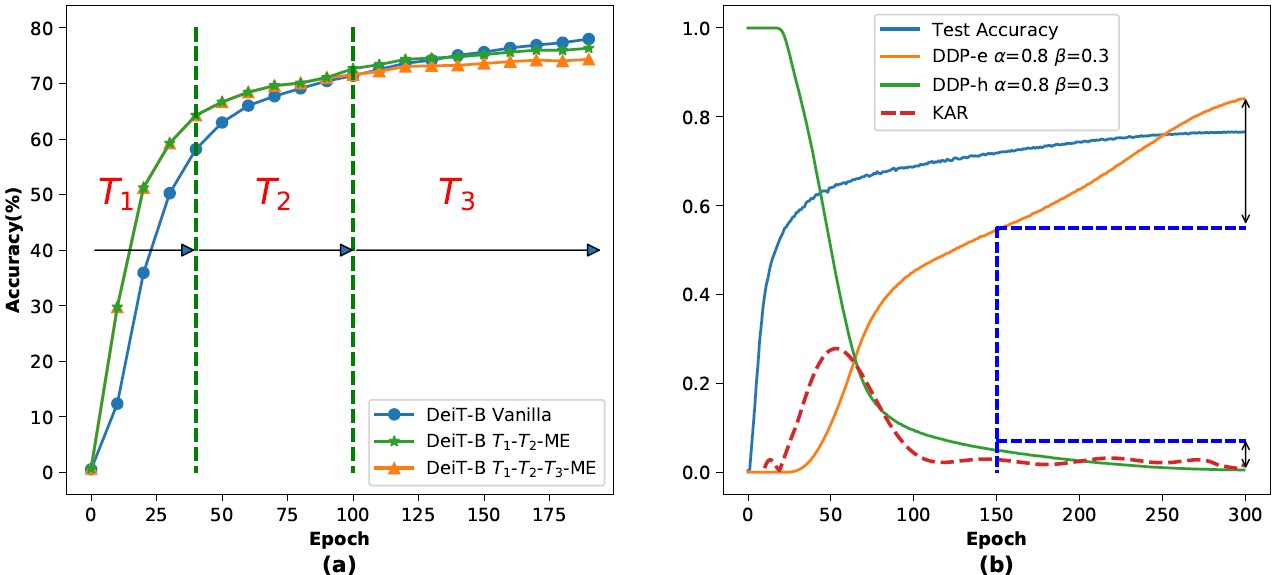}
    \caption{(a) Configuring the easy samples throughout the ($T_1$, $T_2$) or ($T_1$, $T_2$, $T_3$) stages for DeiT-B. (b) Another view to explain the phenomenon of overfitting, \ie, a rapid increase of $\mathrm{DDP}_{t,e}$ in the late $T_3$ period does not bring additional benefits to the performance (herein we adopt several lighter data augmentation strategies instead of AutoAugment \cite{cubuk2018autoaugment} to show the overfitting phenomenon more intuitively on DeiT-S).}
    \label{fig:T1_T2_ME}
\end{wrapfigure}
\paragraph{Growth Period} At the beginning of this period, the model already has the ability to challenge these more complicated examples, so it now focuses on exploring new knowledge from moderate data space $\mathcal{D}_{t, m}$. 
The performance of ViTs with different capacities is comparatively different in this period. For example, the KAR of DeiT-T is much lower than that of DeiT-S and DeiT-B under the same data settings as shown in Figure~\ref{fig:deitperiod}. It means that smaller capacity networks need more time to assimilate and conform the knowledge from $\mathcal{D}_{t, m}$. 

It is surprised to find that the first two periods have a significant impact on the final performance,  especially for large models. As shown in Figure.\ref{fig:T1_T2_ME}(a), if we configure easy samples throughout these periods on DeiT-B, the model can be fitted quicker in these stages while reducing the generalization ability to some extent, resulting in the degradation of the final performance as illustrated in Table.\ref{tab:evolu_invest}. Therefore, this period plays a pivotal role in the final formation of a strong model, and over-easy or -hard sample distributions are not conducive for model training. Therefore, in this period, we propose to leverage a rapid-growing strategy to dynamically control the data space to follow a distribution from easy to hard. 


\paragraph{Exploration Period} The $T_3$ evolution period, occupying most of the training time, is a stressful yet meaningful time for ViTs. During this period, the KAR falls back into a lower level as shown in Figure~\ref{fig:deitperiod}. It means that the network encountered greater learning resistance during this period, \ie, the network tries to transfer the assimilation target to these harder examples (\eg, the examples that occupy the middle ground between $\mathcal{D}_{t, m}$ and $\mathcal{D}_{t, h}$, as well as hard examples in $\mathcal{D}_{t, h}$), and gradually obtains the ability to induct high-level semantic knowledge. 

As can be seen from Figure.\ref{fig:deitperiod}, as the model is trained, the increase of $\mathrm{DDP}_{t,e}$ and decrease of $\mathrm{DDP}_{t,h}$ show an approximately symmetric nature, which means that the model gradually learns more discriminative concepts from hard samples, thereby continuously improving the confidence of easy and moderate samples. In addition, we find that models with large capacity have faster knowledge assimilation speed and more substantial knowledge reserves in this period, thus forming stronger classifiers.
For example, after training for 250 epochs, DeiT-B has less than about 10\% hard samples left, while DeiT-T is still a staggering about 30\%. This also means that DeiT-B will face the dilemma of knowledge exhaustion in the limited data space.


Intuitively, a sufficiently long $T_3$ period allows the model to absorb as much new knowledge as possible from a hard sample space, to consistently improve performance. 
However, with long-term training, a natural problem comes up: the network has difficulty continuing to absorb new knowledge from the limited data space, whereby prompting it to focus on those samples that have already been assimilated, which is commonly known as the overfitting phenomenon. Exactly, as can be seen in Figure.\ref{fig:T1_T2_ME}(b), the continuous decrease of $\mathrm{DDP}_{t,h}$ is accompanied by a rapid increase of $\mathrm{DDP}_{t,e}$ in late $T_3$ period, but does not bring additional benefits to the performance. The explanation is that ``effective'' examples for the network are running out. 
\begin{table}
  \caption{\textbf{Patch-level overfitting phenomenon of ViTs.} The results below refer to predictions on the training set of ImageNet1K (refer to Appendix for Swin Transformer). Wherein Top-k represents the top k image patches with the highest response, here we selected through the \textit{self-attention scores of the last layer of Transformer}, and $p_k$ represents the mean of all the predicted maximum softmax scores in $\mathcal{D}$.
}
  \label{tab:overfit}
  \centering
   \resizebox{1.\linewidth}{!}{
  \begin{tabular}{l|ll|ll|ll|ll}
    \toprule
    \multirow{2}{*}{Top-k} & \multicolumn{2}{c}{DeiT-S} &  \multicolumn{2}{c}{DeiT-B}          &  \multicolumn{2}{c}{Swin-S} & \multicolumn{2}{c}{Swin-B}  \\
    \cmidrule(r){2-9} 
     & $p_k$  & Top-1 & $p_k$  & Top-1 & $p_k$  & Top-1 & $p_k$  & Top-1 \\

    \midrule
      0 & 0.67 & 77.8 & 0.75 & 85.5  & 0.631 & 80.2 &  0.631 & 82.0  \\
      1 & $0.67_{\textcolor{red}{-0.00}}$ & $77.5_{\textcolor{red}{-0.3}}$ &
      $0.75_{\textcolor{red}{-0.00}}$ & $85.5_{\textcolor{red}{-0.0}}$ & $0.631_{\textcolor{red}{-0.000}}$ & $80.0_{\textcolor{red}{-0.2}}$ &  $0.631_{\textcolor{red}{-0.000}}$ & $81.9_{\textcolor{red}{-0.1}}$ \\
      5 & $0.66_{\textcolor{red}{-0.01}}$ & $76.0_{\textcolor{red}{-1.8}}$ & 
      $0.75_{\textcolor{red}{-0.00}}$ & $85.2_{\textcolor{red}{-0.3}}$ & $0.630_{\textcolor{red}{-0.001}}$ & $79.7_{\textcolor{red}{-0.5}}$  &  $0.631_{\textcolor{red}{-0.000}}$ & $81.5_{\textcolor{red}{-0.5}}$    \\
      10 & $0.64_{\textcolor{red}{-0.03}}$ & $74.3_{\textcolor{red}{-3.5}}$ &
      $0.74_{\textcolor{red}{-0.01}}$ & $84.9_{\textcolor{red}{-0.6}}$ & $0.627_{\textcolor{red}{-0.004}}$ & $78.9_{\textcolor{red}{-1.3}}$  &  $0.630_{\textcolor{red}{-0.001}}$ & $81.0_{\textcolor{red}{-1.0}}$  \\
    
    \bottomrule
  \end{tabular}
  }
\end{table}
\paragraph{Overfitting Investigation.}
In fact, we observe that ViTs show a patch-level overfitting phenomenon in the late stage, suggesting that several image patches dominate the semantic content of the entire image. To dive into this phenomenon, we analyze the effect on the average accuracy and softmax score on the training set of ImageNet1K when erased the largest top-k patches in the image that the model ``cares about'' the most. More specifically, we use the parameters with the highest accuracy on the test set to make predictions on the training set. Similar to the inference stage, the predicted samples are first resized to $256\times256$, and then center-cropped into $224\times224$. 
As illustrated in Table~\ref{tab:overfit}, the average maximum softmax score, $p_k=\frac{1}{K}\sum_{i}^{K}{p_{k_i}}, ~K = \vert\mathcal{D}\vert$, is almost unchanged if the input is fed into ViTs after erasing a few most sensitive patch tokens, but the accuracy degrades significantly. This phenomenon reveals that the final prediction of the model is determined by only a few patches in the image, resulting in a poor generalization ability of ViTs.

Based on the above observations, we propose a dynamic MixUp scheme to generate more ``effective'' examples at the different training stages. 
Meanwhile, in order to alleviate the overfitting phenomenon of models in the later period of $T_3$, we propose a PatchErasing strategy to increase the knowledge capacity of the data space. We discuss our method in detail in the following sections.

\section{Methodology}
This paper proposes a heuristic data-centric training framework for Vision Transformers (ViTs), which can be used to tackle many data-centric tasks. Specifically, we present a dynamic and controllable optimization method to dynamically control the distribution ``difficulty'' of training examples based on MixUp~\cite{zhang2017mixup}, to generate enough ``effective'' training examples over the three evolution periods of ViTs optimization. In addition, a new data augmentation strategy, dubbed PatchErasing, is proposed to enrich the knowledge capacity of the training data space.

\subsection{Training Framework}
\label{sec:framework}
\begin{wrapfigure}[12]{r}{0.55\textwidth}
  \centering
    \includegraphics[width=\linewidth]{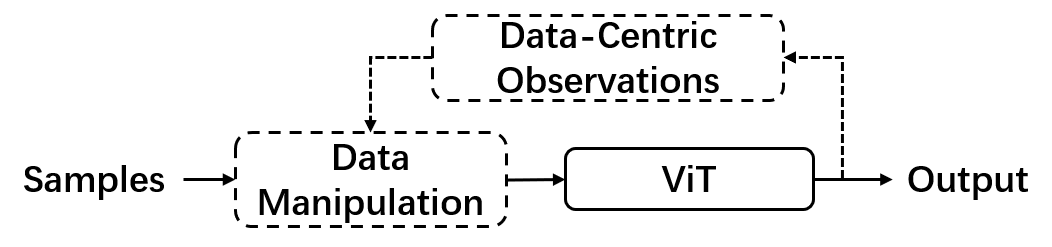}

  \caption{\textbf{The proposed training framework.} During the training, the input data is first processed appropriately according to the data-centric observations derived from the previous iteration, then input to ViT for targeted training.}
  \label{fig:framework}
\end{wrapfigure}
As illustrated in Figure~\ref{fig:framework}, the presented training scheme is mainly divided into two processes, \textit{parameters updating} and \textit{data manipulation}.
During network training, first, we randomly sample a large chunk of data and dynamically control its distribution “difficulty” based on observations of the network state at the previous training moment. Then the tuned data is fed into the network to perform standard forward and backward propagation while recording the predefined measurements of the model at the current moment, such as DDP and KAR as defined above. This new training paradigm can be regarded as data-centered learning based on apriori knowledge. Given the observations on the performance of ViTs at different training stages, we present two approaches, dynamic-based and static-based, to manipulate the ``difficulty'' distribution of training data in this paper. The following sections discuss them in detail.

\subsection{Data Manipulation}
In this section, we first briefly review the definition of MixUp, as our data manipulation methods are based on a dynamic MixUp strategy (one can flexibly replace with other data manipulation methods according to the actual situation), and then introduce two different data manipulation methods and the PatchErasing strategy.
\subsubsection{MixUp.} Generally, MixUp creates new
instances ($\hat{x}$, $\hat{y}$) by randomly taking convex combinations pairs ($x_i$, $y_i$) and ($x_j$, $y_j$), which can be formulated as:
\begin{equation}
    \left\{
\begin{aligned}
&\hat{x} & = & \lambda \cdot x_i+(1-\lambda)\cdot x_j, \\
&\hat{y} & = & \lambda \cdot y_i + (1-\lambda) \cdot y_j.
\end{aligned}
\right.
\end{equation} 
where $\lambda \sim Bata_{[0,1]}(\alpha, \alpha)$ controls the strength of interpolation between sample pairs. Normally, $\alpha$ is a relatively small fixed value, meaning that the distribution ``difficulty'' of examples maintains a standard strength throughout the training.
\textit{Therefore, our goal is to achieve data “difficulty” transformation by dynamically adjusting $\alpha$.} We investigate two distributions of $\alpha$ and evaluate them on networks with different capacities.

\subsubsection{Dynamic-based Distribution.} 
This strategy can adjust the data ``difficulty'' in real-time depending on historical learning conditions. Let $\mathrm{DDP}_{t-1,e}$ and $\mathrm{DDP}_{t-1,h}$ denote the easy and hard examples proportion measured at $t$-1 iteration,
then the $\alpha$ at the $t$-th iteration can be formulated as:
\begin{equation}
    \alpha_t = \frac{1}{2}(\mathrm{DDP}_{t-1,e} + (1-\mathrm{DDP}_{t-1,h}))
\end{equation}
For numerical stability, we use Exponential Moving Average (EMA) to estimate the local mean of the variable $\alpha_t$, so that the update of the $\alpha_t$ is related to the historical value over a period of time:
\begin{equation}
    \bar\alpha_t \leftarrow \tau \bar\alpha_{t-1} + (1-\tau)\alpha_t
\end{equation}
where $\tau$ is momentum, here we set $\tau$=0.9 in all of our experiments. 

\begin{wrapfigure}[20]{r}{0.55\textwidth}
  \centering
    \includegraphics[width=0.9\linewidth]{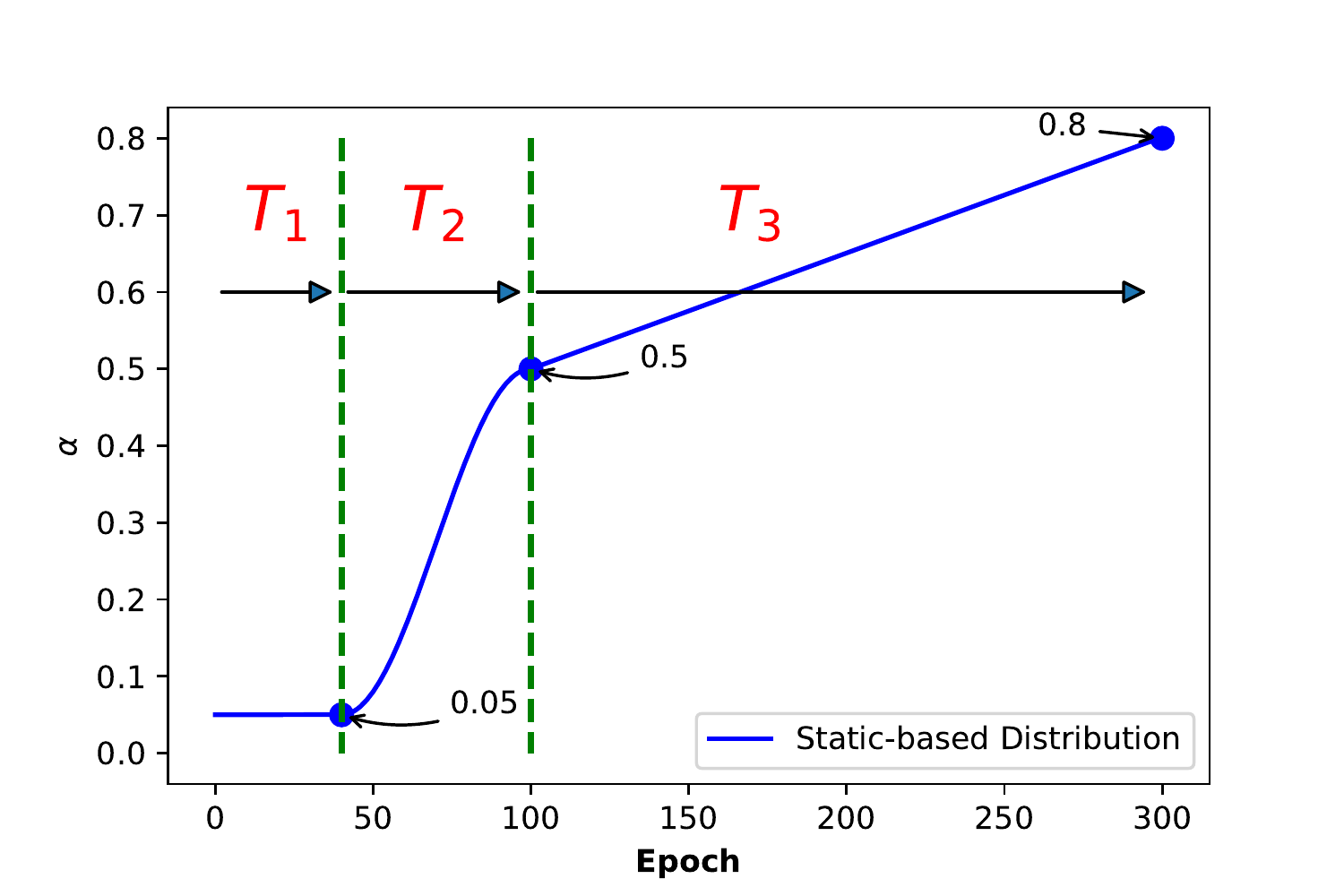}
  \caption{\textbf{The static-based ``difficulty'' distribution.} Wherein the $T_1$ stage utilizes a fixed $\alpha$ of 0.05, the $T_2$ stage rapidly increases $\alpha$ from 0.05 to 0.5 through a cosine-growth function, and the $T_3$ stage slowly increases $\alpha$ from 0.5 to 0.8 through a linear-growth function. }
  \label{fig:scd}
\end{wrapfigure}
\subsubsection{Static-based Distribution.} As shown in Figure.~\ref{fig:scd}, this strategy is employed to manipulate training examples in stages, given the prior that the network enjoys different knowledge assimilation abilities in the three evolution periods as illustrated in Figure.~\ref{fig:deitperiod}. We use a cosine-like growth approach to progressively adjust the $\alpha$ from 0.05 to 0.8 (standard setting in DeiT) during training.

More specifically, In the $T_1$ period, we adopt a relatively low value (about 0.05) to ensure that the ``difficulty'' distribution of training examples is at a relatively low $\alpha$. In the $T_2$ period, we rapidly increase the data ``difficulty'' through a cosine-growing strategy (increasing $\alpha$ from 0.05 to 0.5), given the $T_2$ period is critical for forming a robust model, \ie, over-simple training examples lead to poor generalization, and vice versa, low learning efficiency.
Considering that the $T_3$ period is a long-term process of knowledge exploration, we thus employ a linear-growing strategy (increasing $\alpha$ from 0.5 to 0.8) to progressively improve the data ``difficulty'' in this stage.

\paragraph{PatchErasing.}
PatchErasing is a simple yet effective regularization method that randomly erases a certain number of patches from the image during training. Different from HaS~\cite{singh2018hide} and GridMask~\cite{chen2020gridmask}, PatchErasing is specially designed for ViTs to alleviate its patch-level overfitting phenomenon, \ie, the size of the erasing area is aligned with the partitioned patch inputs for ViT. This is also inspired by MAE~\cite{He_2022_CVPR}, \ie, the image can be reconstructed with only a few patches thanks to the powerful global attention ability of ViTs, which, if unconstrained, also makes the model more sensitive to some local patches during training. Therefore, PatchErasing is proposed to relieve the model from getting trapped in locally optimal solutions. 
Let ($x_i$, $y_i$) and ($x_j$, $y_j$) denote a sample pair, $\gamma_i \in [0, \mu]$ and $\gamma_j \in [0, \mu]$ represent their erasing rate, and $\mathcal{E}(\cdot, \gamma)$ means random erasing function with mask rate $\gamma$. Then the generated new instance can be denoted as: 
\begin{equation} \label{eq:patcherase}
    \left\{
\begin{aligned}
&\hat{x}' & = & \lambda \cdot \mathcal{E}(x_i, \gamma_i) + (1-\lambda)\cdot \mathcal{E}(x_j, \gamma_j), \\
&\hat{y}' & = & (\lambda-\epsilon_i)  \cdot y_i + (1-\lambda-\epsilon_j) \cdot y_j.
\end{aligned}
\right.
\end{equation} 
where $\epsilon_i=\frac{\lambda}{\mu} \times \gamma_i$ and $\epsilon_j=\frac{1-\lambda}{\mu} \times \gamma_j$ are information decay factors for label vectors $y_i$ and $y_j$, \ie, describe the informativeness contained in the instance $\hat{x}'$.
This seems to be ignored by previous work, where they treat instances as 100\% informativeness, even though more than half of the image regions are masked out.
Therefore, in our case, we regularize the corresponding one-hot labels along with the images.

\begin{minipage}[t]{0.45\textwidth}
\captionof{table}{\textbf{Effect of two data manipulation methods for DeiT of different capacities.} Vanilla means traditional training strategy.}
    \label{tab:ded_scd}
    \centering
    \resizebox{1.\linewidth}{!}{
    \begin{tabular}{c|ccc}
    \toprule
    Strategy & DeiT-T & DeiT-S & DeiT-B \\
    \midrule
    Vanilla & 72.2 & 79.8 & 81.8\\
     dynamic-based    & \textbf{72.6} & \textbf{80.3} & 82.2\\
     static-based    & 72.4 & 80.2 & \textbf{82.3}\\
    \bottomrule
    \end{tabular}
    }
\end{minipage}
\hfill%
\begin{minipage}[t]{0.5\textwidth}
\captionof{table}{\textbf{Effects of erasing rate for PatchErasing}. Wherein RE and PE represent vanilla RandomErasing and proposed PatchErasing, respectively. Typically, the $\gamma \sim Unif([0, \mu])$ is a random variables with $\mu \geq 0$.}
    \label{tab:patch_erasing}
    \centering
    \resizebox{1.\linewidth}{!}{
    \begin{tabular}{ccc|ccc}
    \toprule
    RE & PE & Mask Rate ($\gamma$) & DeiT-T & DeiT-S & DeiT-B \\
    \midrule
   \CheckmarkBold & & - & 72.2 & 79.8 & 81.8\\
   & \CheckmarkBold & 0\% $\sim$ 10\%  & \textbf{72.5} & 80.0 & 81.9\\
   \rowcolor[gray]{0.8} & \CheckmarkBold & 0\% $\sim$  50\%  & 72.4 & \textbf{80.1} & 82.1\\
   & \CheckmarkBold & 0\% $\sim$  100\%  & 71.8 & 79.9 & \textbf{82.2}\\
    \bottomrule
    \end{tabular}
    }
\end{minipage}
\section{Experiments}
We evaluate the proposed data-centric training framework and data augmentation strategy on two representatives ViT architectures, DeiT~\cite{touvron2021training} and Swin Transformer~\cite{liu2021swin}. DeiT is a global ViT and Swin Transformer is a local ViT. (We also transfer it to CNNs to show its generality and scalability in Appendix.) 

\paragraph{Settings.} All evaluations are conducted on the ImageNet1K ~\cite{deng2009imagenet} dataset. Our experiments are conducted with Pytorch ~\cite{paszke2019pytorch} and timm ~\cite{rw2019timm} on Tesla V100-32GB GPUs. No extra training images or parameters are introduced. During training for DeiT and Swin Transformer, their official hyper-parameter configurations are followed, including the total 300 epochs of training. We also leverage the same augmentation and regularization strategies as DeiT and Swin Transformer. All models are trained/evaluated on $224 \times 224$ resolution unless otherwise specified. It is worth noting that the proposed techniques are only employed in the training phase.

\subsection{Ablation Study}
In this section, the DeiT family is used for research. All settings are in accordance with the official configuration unless otherwise specified.
\begin{figure}[t]
  \centering
     \begin{subfigure}{0.30\textwidth}
    \includegraphics[width=\linewidth]{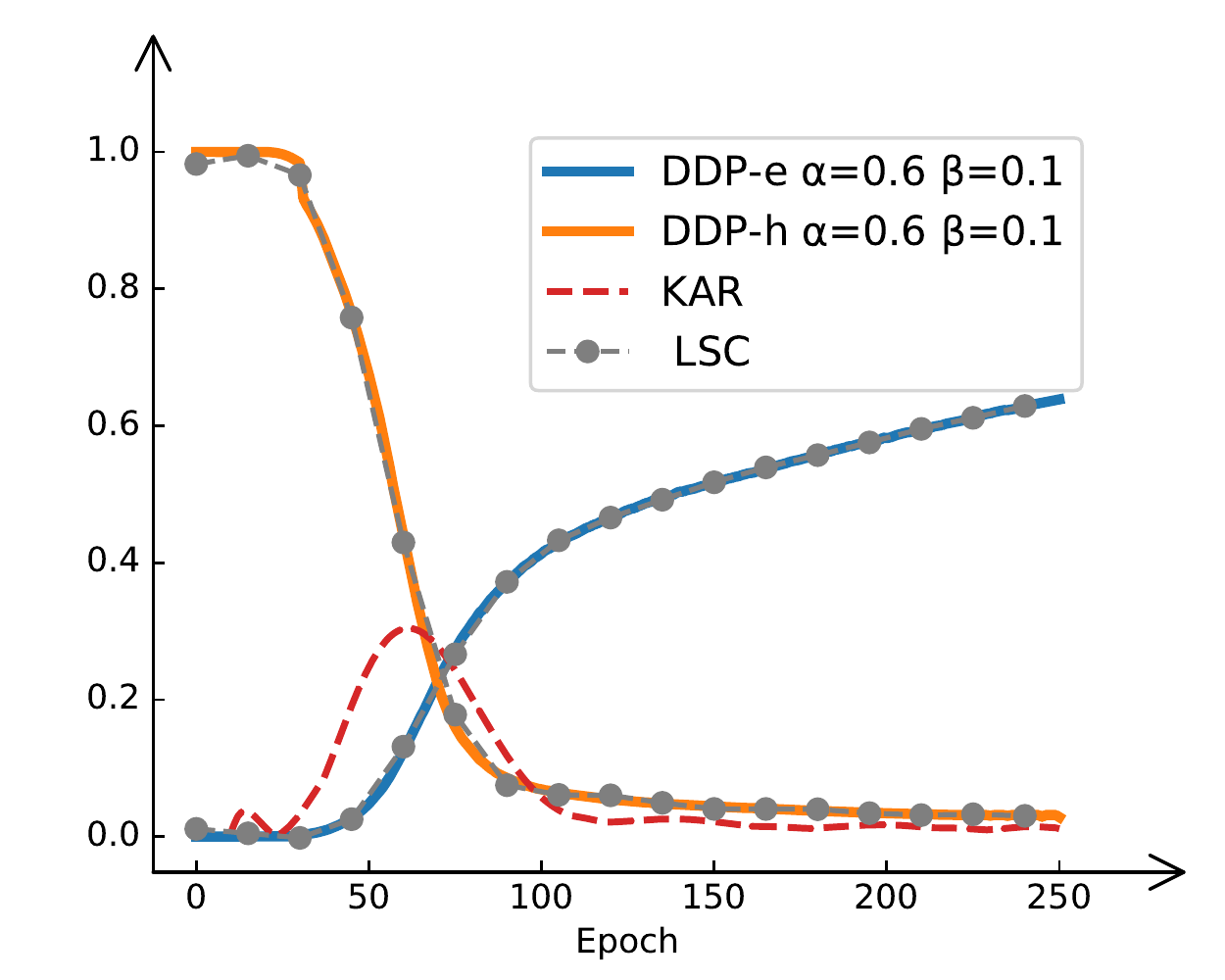}
    \caption{DeiT-S: $\alpha$=0.6, $\beta$=0.1.}
    \label{subfig:swin_small}
    \end{subfigure}
    \hfill
    \begin{subfigure}{0.30\textwidth}
    \includegraphics[width=\linewidth]{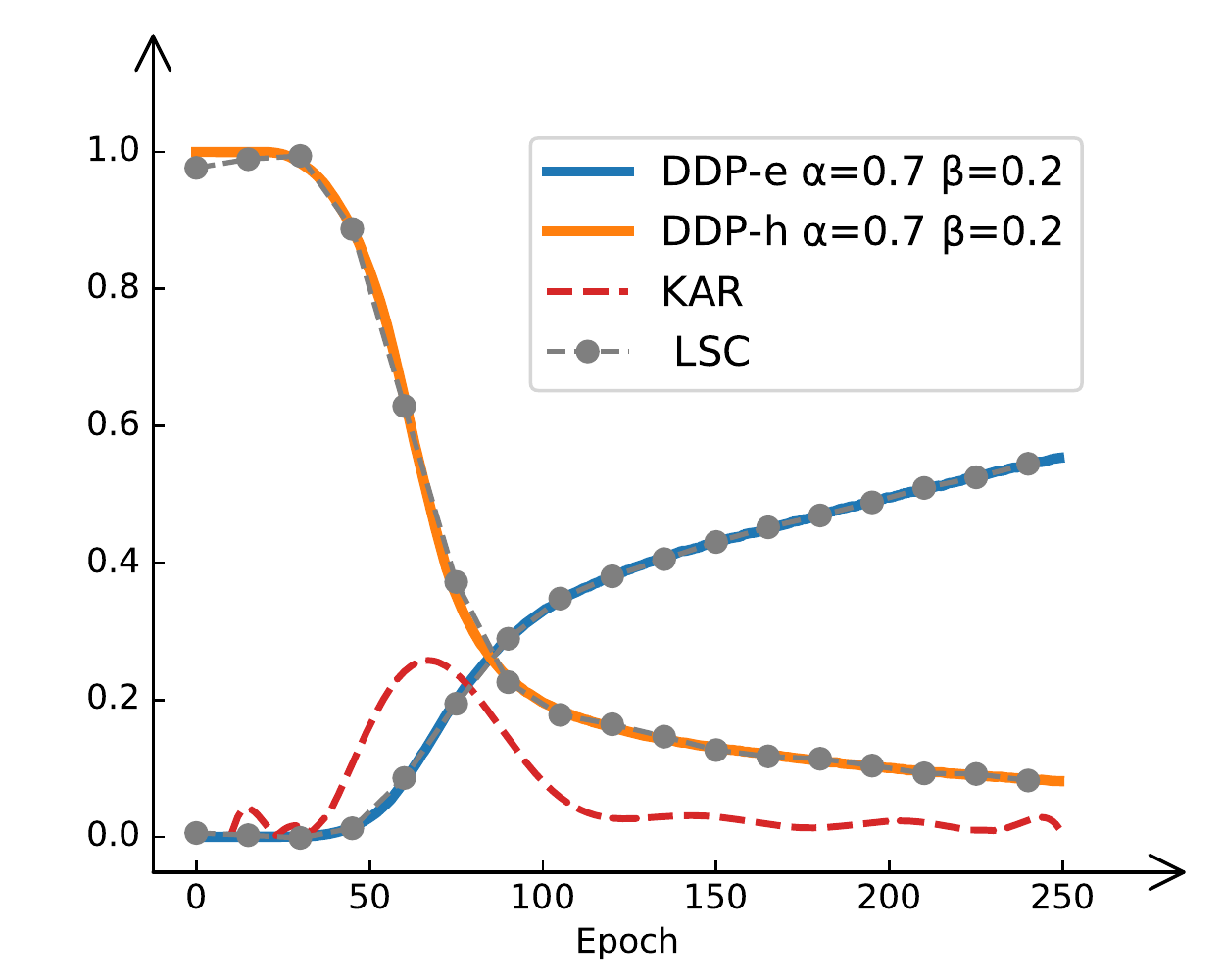}
    \caption{DeiT-S: $\alpha$=0.7, $\beta$=0.2.}
    \label{subfig:deit_tiny}
    \end{subfigure}
    \hfill
    \begin{subfigure}{0.30\textwidth}
    \includegraphics[width=\linewidth]{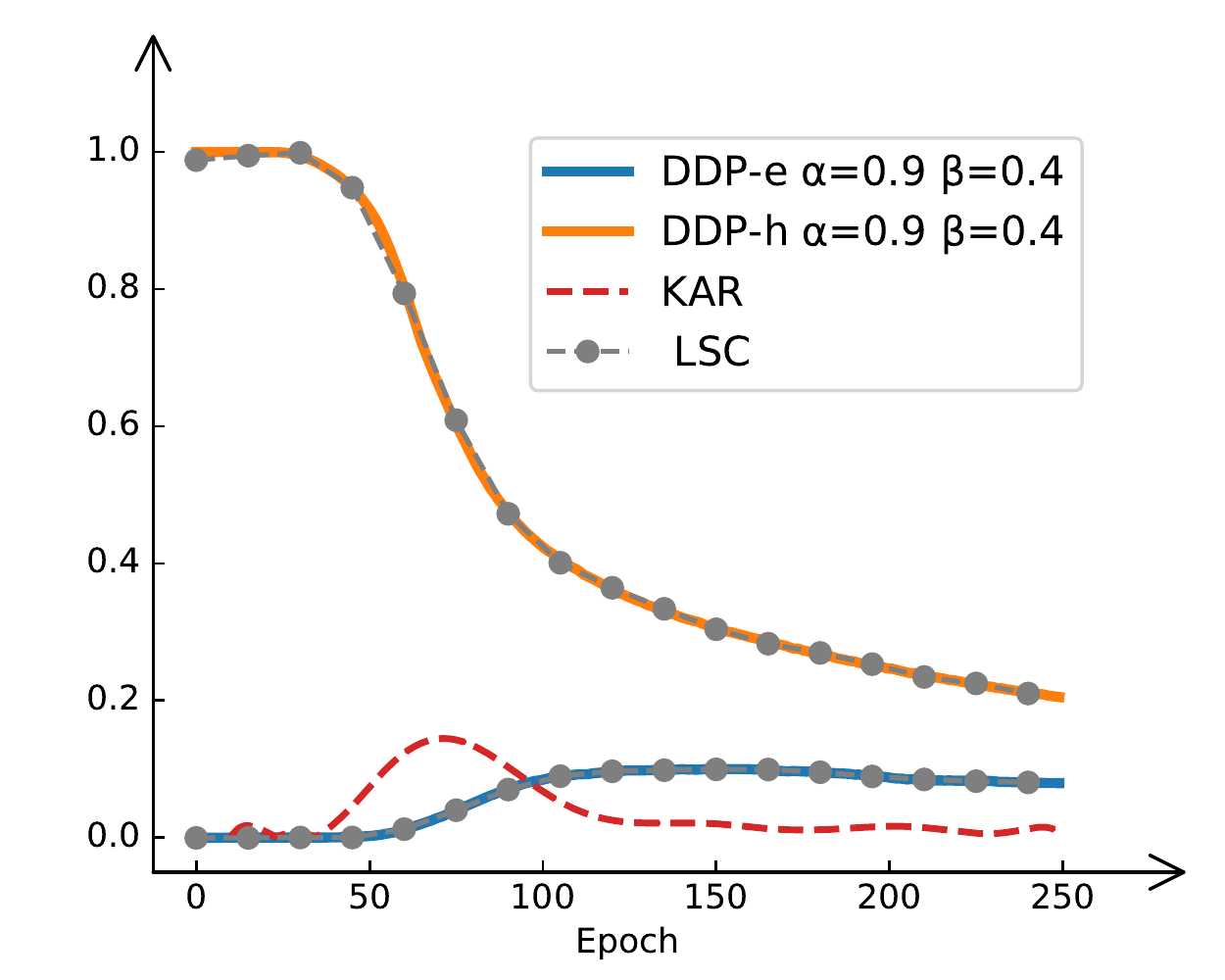}
    \caption{DeiT-S: $\alpha$=0.9, $\beta$=0.4.}
    \label{subfig:deit_base}
    \end{subfigure}

  \caption{\textbf{The evolution period of DeiT-S under different thresholds $\alpha$ and $\beta$.} For more evolution curves please refer to Appendix.}
  \label{fig:deit_period}
\end{figure}

\begin{minipage}[c]{0.55\textwidth}
\captionof{table}{\textbf{Comparison on different backbones}. \#Res represents resolutions used in training / validating, and * indicates our reproduced results.}
  \label{tab:main}
\centering
  \small
  \resizebox{1.\linewidth}{!}{
  \begin{tabular}{lllllll}
    \toprule
    \multirow{2}{*}{Model}& \multirow{2}{*}{Params} & \multirow{2}{*}{\#Res} & \multicolumn{2}{c}{Baseline} & \multicolumn{2}{c}{Ours} \\
    \cmidrule(r){4-7}
    &  &   & Top-1  & Top-5 & Top-1  & Top-5\\
    \midrule
    DeiT-T     & 5M   & $224^2$ &  72.2  &   91.1  &  $72.8_{\textcolor{blue}{+0.6}}$ & $91.4_{\textcolor{blue}{+0.3}}$\\
    DeiT-S     & 22M  & $224^2$ &  79.8  &   95.0  &
    $80.5_{\textcolor{blue}{+0.7}}$ & $95.2_{\textcolor{blue}{+0.2}}$ \\
    DeiT-B     & 86M  & $224^2$ &  81.8  &   95.6   &
    $82.4_{\textcolor{blue}{+0.6}}$ & $95.8_{\textcolor{blue}{+0.2}}$ \\
    DeiT-B$\uparrow 384$     & 86M  & $384^2$ &  82.9*  &  96.2* &
    $83.4_{\textcolor{blue}{+0.5}}$ & $96.3_{\textcolor{blue}{+0.1}}$ \\
    \midrule
    Swin-T     & 28M   & $224^2$ &  81.3  &  95.5   &
    $81.8_{\textcolor{blue}{+0.5}}$ & $95.7_{\textcolor{blue}{+0.2}}$\\
    Swin-S     & 50M   & $224^2$ &  83.0  &  96.2   &
    $83.5_{\textcolor{blue}{+0.5}}$ & $96.3_{\textcolor{blue}{+0.1}}$\\
    Swin-B     & 88M   & $224^2$ &  83.5  &  96.5   &
    $84.0_{\textcolor{blue}{+0.5}}$ & $96.7_{\textcolor{blue}{+0.2}}$\\
    Swin-B$\uparrow 384$     & 88M   & $384^2$ &  84.5  &  97.0 & 
    $84.9_{\textcolor{blue}{+0.4}}$ & $97.2_{\textcolor{blue}{+0.2}}$\\
    \bottomrule
  \end{tabular}
  }
\end{minipage}%
 \hfill%
\begin{minipage}[c]{0.4\textwidth}
\centering
    \includegraphics[width=0.8\linewidth]{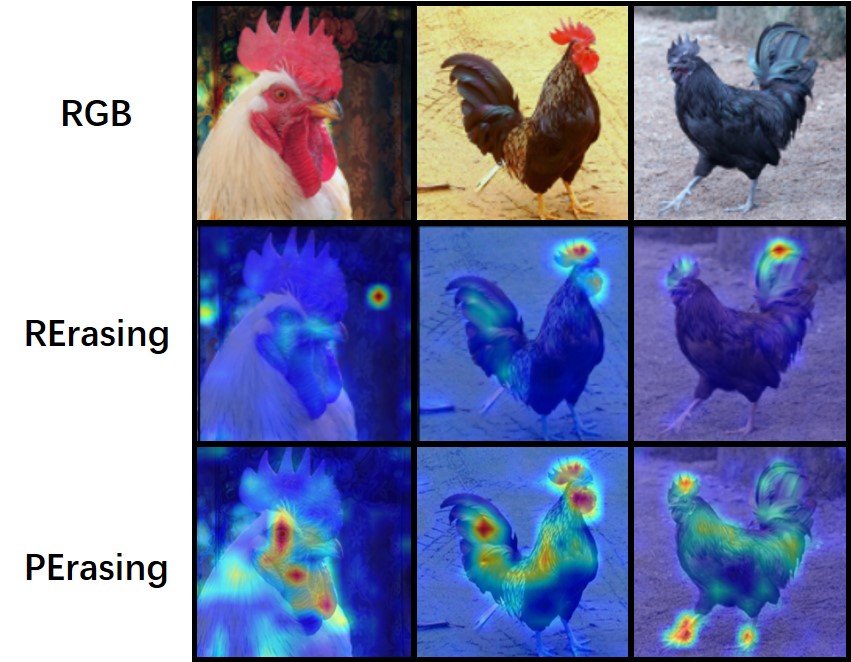}
    \captionof{figure}{Class Activation Map (CAM) of vanilla random erasing strategy (RErasing) and proposed patch erasing strategy (PErasing). Refer to the Appendix for more.}
    \label{fig:visual}
\end{minipage}
\paragraph{Effect of two Data Manipulate Strategies.} 
As can be seen from Table.~\ref{tab:ded_scd}, both two data ``difficulty'' manipulation methods can consistently improve performance (DeiT-T: \textcolor{blue}{+0.4}, DeiT-S: \textcolor{blue}{+0.5}, DeiT-B: \textcolor{blue}{+0.5}), but they maintain their own characteristics. Concretely, the small-scale networks, \eg, DeiT-T and DeiT-S, enjoy the dynamic-based manipulation strategy, which means it is necessary to allocate ``effective'' samples more flexibly in combination with their actual learning ability. However, for large-scale networks, \eg, DeiT-B, it is advisable to rapidly increase the sample ``difficulty'' at the $T_2$ evolution period to avoid it falling into a local optimum, thereby resulting in decreasing the generalization ability of models. Based on the above observations, during training, we employ the static-based manipulation method to control the ``difficulty'' distribution of training data for large-scale networks (\eg, DeiT-B and Swin-B), while the dynamic-based manipulation method is employed for small-scale networks (\eg, DeiT-T, DeiT-S, Swin-T, and Swin-S).

\paragraph{Impact of Erasing Rate for PatchErasing.}
In this paragraph, we investigate the impact of mask rate for PatchErasing. As can be seen from Table.~\ref{tab:patch_erasing}, the mask rate has a significant impact on performance. We find that the large model prefers stronger regularization (promote the performance $\textcolor{blue}{+0.4}$ when $\gamma \in Unif([0, 1])$), which may be because large models are more prone to overfitting. Taking into account the performance gains both on small and large models, we finally choose the hyperparameter $\mu$=50\% in all of our experiments.

\paragraph{Evolution Period Under Different Thresholds.} 
As shown in Figure.~\ref{fig:deit_period}, even under different thresholds $\alpha$ and $\beta$, the evolution period of the network still follows three stages, and the time they experience in these stages is basically the same, which further demonstrates the existence of these evolution periods during training. 
Nevertheless, under a stricter threshold settings, the dynamic data proportions of easy and hard samples are lower. For example, when training to 250 epoch, the $\mathrm{DDP}_{250, e} \leq 20\%$ for $\alpha$=0.9, and the $\mathrm{DDP}_{250, h}\leq 20\%$ for $\beta$=0.1.

\paragraph{Attention Map Visualization}
According to the observation that ViTs may be suffering from patch-level overfitting, we visually show that PatchErasing can alleviate this problem to some extent. As shown in Figure.~\ref{fig:visual}, compared with the vanilla erasing strategy, PatchErasing can force models to focus on more regions of the object rather than just a few locally significant areas, thereby making the models more robust.

\subsection{Main Experimental Results}
We compare the models trained by the proposed training framework with baseline models in Table~\ref{tab:main}. The training strategy we adopt is to dynamically control the ``difficulty'' of training examples at different evolution periods. Furthermore, PatchErasing is introduced in the $T_3$ learning period to enlarge the knowledge capacity of the data space.
As can be seen in Table~\ref{tab:main}, for different model sizes, our proposed method consistently boosts DeiT and Swin Transformer, while almost no extra time overhead is introduced during training.
In particular, our method improves the average performance of DeiT by $\geq0.5\%$, and Swin Transformer by $\geq 0.4\%$, which demonstrates the effectiveness of the proposed technologies in global-based and local-based ViTs. 

\section{Conclusions}
In this paper, we thoroughly investigate the training process of ViTs and divide it into three stages. A data-centric ViT training framework is proposed to measure and generate  ``effective'' training examples. 
In addition, a patch-level erasing strategy is proposed to further enlarge the number of “effective” samples and relieve the overfitting problem in the late training stage. We hope our work can inspire more research works on data-centric neural network training.  

\section*{Acknowledgments}
This work was supported by the National Key Research and Development Plan under Grant 2020YFC2003901, 
the External cooperation key project of Chinese Academy Sciences  173211KYSB20200002, the Chinese National Natural Science Foundation Projects 61876179 and 61961160704, the Science and Technology Development Fund of Macau (0025/2019/AKP, 0008/2019/A1, 0010/2019/AFJ, 0004/2020/A1 0070/2020/AMJ), Guangdong Provincial Key R\&D Programme: 2019B010148001, the InnoHK program, and the open Research Projects of Zhejiang Lab No. 2021PF0AB01, the Alibaba Group through Alibaba Research Intern Program.

{\small
\bibliographystyle{abbrv}
\bibliography{ref}

\begin{thebibliography}{10}

\bibitem{balestriero2022effects}
R.~Balestriero, L.~Bottou, and Y.~LeCun.
\newblock The effects of regularization and data augmentation are class
  dependent.
\newblock {\em arXiv preprint arXiv:2204.03632}, 2022.

\bibitem{bengio2009curriculum}
Y.~Bengio, J.~Louradour, R.~Collobert, and J.~Weston.
\newblock Curriculum learning.
\newblock In {\em Proceedings of the 26th annual international conference on
  machine learning}, pages 41--48, 2009.

\bibitem{chen2021psvit}
B.~Chen, P.~Li, B.~Li, C.~Li, L.~Bai, C.~Lin, M.~Sun, J.~Yan, and W.~Ouyang.
\newblock Psvit: Better vision transformer via token pooling and attention
  sharing.
\newblock {\em arXiv preprint arXiv:2108.03428}, 2021.

\bibitem{chen2021crossvit}
C.-F. Chen, Q.~Fan, and R.~Panda.
\newblock Crossvit: Cross-attention multi-scale vision transformer for image
  classification.
\newblock {\em arXiv preprint arXiv:2103.14899}, 2021.

\bibitem{chen2020gridmask}
P.~Chen, S.~Liu, H.~Zhao, and J.~Jia.
\newblock Gridmask data augmentation.
\newblock {\em arXiv preprint arXiv:2001.04086}, 2020.

\bibitem{Zhengsu21}
Z.~Chen, L.~Xie, J.~Niu, X.~Liu, L.~Wei, and Q.~Tian.
\newblock Visformer: The vision-friendly transformer.
\newblock {\em arXiv preprint arXiv:2104.12533}, 2021.

\bibitem{cubuk2018autoaugment}
E.~D. Cubuk, B.~Zoph, D.~Mane, V.~Vasudevan, and Q.~V. Le.
\newblock Autoaugment: Learning augmentation policies from data.
\newblock {\em arXiv preprint arXiv:1805.09501}, 2018.

\bibitem{cubuk2020randaugment}
E.~D. Cubuk, B.~Zoph, J.~Shlens, and Q.~V. Le.
\newblock Randaugment: Practical automated data augmentation with a reduced
  search space.
\newblock In {\em Proceedings of the IEEE/CVF Conference on Computer Vision and
  Pattern Recognition Workshops}, pages 702--703, 2020.

\bibitem{d2021convit}
S.~d'Ascoli, H.~Touvron, M.~Leavitt, A.~Morcos, G.~Biroli, and L.~Sagun.
\newblock Convit: Improving vision transformers with soft convolutional
  inductive biases.
\newblock {\em arXiv preprint arXiv:2103.10697}, 2021.

\bibitem{deng2009imagenet}
J.~Deng, W.~Dong, R.~Socher, L.-J. Li, K.~Li, and L.~Fei-Fei.
\newblock Imagenet: A large-scale hierarchical image database.
\newblock In {\em 2009 IEEE conference on computer vision and pattern
  recognition}, pages 248--255. Ieee, 2009.

\bibitem{cswin}
X.~Dong, J.~Bao, D.~Chen, W.~Zhang, N.~Yu, L.~Yuan, D.~Chen, and B.~Guo.
\newblock Cswin transformer: A general vision transformer backbone with
  cross-shaped windows.
\newblock {\em arXiv preprint arXiv:2107.00652}, 2021.

\bibitem{dosovitskiy2020image}
A.~Dosovitskiy, L.~Beyer, A.~Kolesnikov, D.~Weissenborn, X.~Zhai,
  T.~Unterthiner, M.~Dehghani, M.~Minderer, G.~Heigold, S.~Gelly, et~al.
\newblock An image is worth 16x16 words: Transformers for image recognition at
  scale.
\newblock {\em arXiv preprint arXiv:2010.11929}, 2020.

\bibitem{el2021xcit}
A.~El-Nouby, H.~Touvron, M.~Caron, P.~Bojanowski, M.~Douze, A.~Joulin,
  I.~Laptev, N.~Neverova, G.~Synnaeve, J.~Verbeek, et~al.
\newblock Xcit: Cross-covariance image transformers.
\newblock {\em arXiv preprint arXiv:2106.09681}, 2021.

\bibitem{fan2021multiscale}
H.~Fan, B.~Xiong, K.~Mangalam, Y.~Li, Z.~Yan, J.~Malik, and C.~Feichtenhofer.
\newblock Multiscale vision transformers.
\newblock {\em arXiv preprint arXiv:2104.11227}, 2021.

\bibitem{gong2021improve}
C.~Gong, D.~Wang, M.~Li, V.~Chandra, and Q.~Liu.
\newblock Improve vision transformers training by suppressing over-smoothing.
\newblock {\em arXiv preprint arXiv:2104.12753}, 2021.

\bibitem{graham2021levit}
B.~Graham, A.~El-Nouby, H.~Touvron, P.~Stock, A.~Joulin, H.~J{\'e}gou, and
  M.~Douze.
\newblock Levit: a vision transformer in convnet's clothing for faster
  inference.
\newblock {\em arXiv preprint arXiv:2104.01136}, 2021.

\bibitem{graves2017automated}
A.~Graves, M.~G. Bellemare, J.~Menick, R.~Munos, and K.~Kavukcuoglu.
\newblock Automated curriculum learning for neural networks.
\newblock In {\em international conference on machine learning}, pages
  1311--1320. PMLR, 2017.

\bibitem{guo2021cmt}
J.~Guo, K.~Han, H.~Wu, C.~Xu, Y.~Tang, C.~Xu, and Y.~Wang.
\newblock Cmt: Convolutional neural networks meet vision transformers.
\newblock {\em arXiv preprint arXiv:2107.06263}, 2021.

\bibitem{han2021transformer}
K.~Han, A.~Xiao, E.~Wu, J.~Guo, C.~Xu, and Y.~Wang.
\newblock Transformer in transformer.
\newblock {\em arXiv preprint arXiv:2103.00112}, 2021.

\bibitem{He_2022_CVPR}
K.~He, X.~Chen, S.~Xie, Y.~Li, P.~Doll\'ar, and R.~Girshick.
\newblock Masked autoencoders are scalable vision learners.
\newblock In {\em Proceedings of the IEEE/CVF Conference on Computer Vision and
  Pattern Recognition (CVPR)}, pages 16000--16009, June 2022.

\bibitem{he2016deep}
K.~He, X.~Zhang, S.~Ren, and J.~Sun.
\newblock Deep residual learning for image recognition.
\newblock In {\em Proceedings of the IEEE conference on computer vision and
  pattern recognition}, pages 770--778, 2016.

\bibitem{heo2021rethinking}
B.~Heo, S.~Yun, D.~Han, S.~Chun, J.~Choe, and S.~J. Oh.
\newblock Rethinking spatial dimensions of vision transformers.
\newblock {\em arXiv preprint arXiv:2103.16302}, 2021.

\bibitem{huang2021shuffle}
Z.~Huang, Y.~Ben, G.~Luo, P.~Cheng, G.~Yu, and B.~Fu.
\newblock Shuffle transformer: Rethinking spatial shuffle for vision
  transformer.
\newblock {\em arXiv preprint arXiv:2106.03650}, 2021.

\bibitem{jiang2021token}
Z.~Jiang, Q.~Hou, L.~Yuan, D.~Zhou, X.~Jin, A.~Wang, and J.~Feng.
\newblock Token labeling: Training a 85.5\% top-1 accuracy vision transformer
  with 56m parameters on imagenet.
\newblock {\em arXiv preprint arXiv:2104.10858}, 2021.

\bibitem{khan2021transformers}
S.~Khan, M.~Naseer, M.~Hayat, S.~W. Zamir, F.~S. Khan, and M.~Shah.
\newblock Transformers in vision: A survey.
\newblock {\em ACM Computing Surveys (CSUR)}, 2021.

\bibitem{li2021localvit}
Y.~Li, K.~Zhang, J.~Cao, R.~Timofte, and L.~Van~Gool.
\newblock Localvit: Bringing locality to vision transformers.
\newblock {\em arXiv preprint arXiv:2104.05707}, 2021.

\bibitem{liu2021swin}
Z.~Liu, Y.~Lin, Y.~Cao, H.~Hu, Y.~Wei, Z.~Zhang, S.~Lin, and B.~Guo.
\newblock Swin transformer: Hierarchical vision transformer using shifted
  windows.
\newblock In {\em Proceedings of the IEEE/CVF International Conference on
  Computer Vision}, pages 10012--10022, 2021.

\bibitem{mobilevit}
S.~Mehta and M.~Rastegari.
\newblock Mobilevit: Light-weight, general-purpose, and mobile-friendly vision
  transformer.
\newblock {\em arXiv preprint arXiv:2110.02178}, 2021.

\bibitem{paszke2019pytorch}
A.~Paszke, S.~Gross, F.~Massa, A.~Lerer, J.~Bradbury, G.~Chanan, T.~Killeen,
  Z.~Lin, N.~Gimelshein, L.~Antiga, et~al.
\newblock Pytorch: An imperative style, high-performance deep learning library.
\newblock {\em Advances in neural information processing systems}, 32, 2019.

\bibitem{rao2021dynamicvit}
Y.~Rao, W.~Zhao, B.~Liu, J.~Lu, J.~Zhou, and C.-J. Hsieh.
\newblock Dynamicvit: Efficient vision transformers with dynamic token
  sparsification.
\newblock {\em arXiv preprint arXiv:2106.02034}, 2021.

\bibitem{shorten2019survey}
C.~Shorten and T.~M. Khoshgoftaar.
\newblock A survey on image data augmentation for deep learning.
\newblock {\em Journal of big data}, 6(1):1--48, 2019.

\bibitem{singh2018hide}
K.~K. Singh, H.~Yu, A.~Sarmasi, G.~Pradeep, and Y.~J. Lee.
\newblock Hide-and-seek: A data augmentation technique for weakly-supervised
  localization and beyond.
\newblock {\em arXiv preprint arXiv:1811.02545}, 2018.

\bibitem{touvron2021training}
H.~Touvron, M.~Cord, M.~Douze, F.~Massa, A.~Sablayrolles, and H.~J{\'e}gou.
\newblock Training data-efficient image transformers \& distillation through
  attention.
\newblock In {\em International Conference on Machine Learning}, pages
  10347--10357, 2021.

\bibitem{touvron2021going}
H.~Touvron, M.~Cord, A.~Sablayrolles, G.~Synnaeve, and H.~J{\'e}gou.
\newblock Going deeper with image transformers.
\newblock {\em arXiv preprint arXiv:2103.17239}, 2021.

\bibitem{wang2021scaled}
P.~Wang, X.~Wang, H.~Luo, J.~Zhou, Z.~Zhou, F.~Wang, H.~Li, and R.~Jin.
\newblock Scaled relu matters for training vision transformers.
\newblock In {\em Proceedings of the AAAI Conference on Artificial Intelligence
  (AAAI)}, 2022.

\bibitem{wang2021kvt}
P.~Wang, X.~Wang, F.~Wang, M.~Lin, S.~Chang, H.~Li, and R.~Jin.
\newblock Kvt: k-nn attention for boosting vision transformers.
\newblock In {\em ECCV}, 2022.

\bibitem{rw2019timm}
R.~Wightman.
\newblock Pytorch image models.
\newblock \url{https://github.com/rwightman/pytorch-image-models}, 2019.

\bibitem{wu2021cvt}
H.~Wu, B.~Xiao, N.~Codella, M.~Liu, X.~Dai, L.~Yuan, and L.~Zhang.
\newblock Cvt: Introducing convolutions to vision transformers.
\newblock {\em arXiv preprint arXiv:2103.15808}, 2021.

\bibitem{xiang2020learning}
L.~Xiang, G.~Ding, and J.~Han.
\newblock Learning from multiple experts: Self-paced knowledge distillation for
  long-tailed classification.
\newblock In {\em European Conference on Computer Vision}, pages 247--263.
  Springer, 2020.

\bibitem{xiao2021early}
T.~Xiao, M.~Singh, E.~Mintun, T.~Darrell, P.~Doll{\'a}r, and R.~Girshick.
\newblock Early convolutions help transformers see better.
\newblock {\em arXiv preprint arXiv:2106.14881}, 2021.

\bibitem{xie2021simmim}
Z.~Xie, Z.~Zhang, Y.~Cao, Y.~Lin, J.~Bao, Z.~Yao, Q.~Dai, and H.~Hu.
\newblock Simmim: A simple framework for masked image modeling.
\newblock {\em arXiv preprint arXiv:2111.09886}, 2021.

\bibitem{xu2022comprehensive}
M.~Xu, S.~Yoon, A.~Fuentes, and D.~S. Park.
\newblock A comprehensive survey of image augmentation techniques for deep
  learning.
\newblock {\em arXiv preprint arXiv:2205.01491}, 2022.

\bibitem{xu2021co}
W.~Xu, Y.~Xu, T.~Chang, and Z.~Tu.
\newblock Co-scale conv-attentional image transformers.
\newblock {\em arXiv preprint arXiv:2104.06399}, 2021.

\bibitem{yuan2021incorporating}
K.~Yuan, S.~Guo, Z.~Liu, A.~Zhou, F.~Yu, and W.~Wu.
\newblock Incorporating convolution designs into visual transformers.
\newblock {\em arXiv preprint arXiv:2103.11816}, 2021.

\bibitem{yuan2021volo}
L.~Yuan, Q.~Hou, Z.~Jiang, J.~Feng, and S.~Yan.
\newblock Volo: Vision outlooker for visual recognition.
\newblock {\em arXiv preprint arXiv:2106.13112}, 2021.

\bibitem{hrformer}
Y.~Yuan, R.~Fu, L.~Huang, W.~Lin, C.~Zhang, X.~Chen, and J.~Wang.
\newblock Hrformer: High-resolution transformer for dense prediction.
\newblock {\em NeurIPS}, 2021.

\bibitem{zhang2017mixup}
H.~Zhang, M.~Cisse, Y.~N. Dauphin, and D.~Lopez-Paz.
\newblock mixup: Beyond empirical risk minimization.
\newblock {\em arXiv preprint arXiv:1710.09412}, 2017.

\bibitem{zhong2020random}
Z.~Zhong, L.~Zheng, G.~Kang, S.~Li, and Y.~Yang.
\newblock Random erasing data augmentation.
\newblock In {\em Proceedings of the AAAI conference on artificial
  intelligence}, volume~34, pages 13001--13008, 2020.

\bibitem{zhou2021deepvit}
D.~Zhou, B.~Kang, X.~Jin, L.~Yang, X.~Lian, Q.~Hou, and J.~Feng.
\newblock Deepvit: Towards deeper vision transformer.
\newblock {\em arXiv preprint arXiv:2103.11886}, 2021.

\bibitem{zhou2021elsa}
J.~Zhou, P.~Wang, F.~Wang, Q.~Liu, H.~Li, and R.~Jin.
\newblock Elsa: Enhanced local self-attention for vision transformer.
\newblock {\em arXiv preprint arXiv:2112.12786}, 2021.

\end{thebibliography}
}

\newpage
\appendix
\section{Appendix}
\subsection{Experiment Results on CNNs} \label{cnns}
\begin{table}[th]
\caption{\textbf{Comparison on ResNet-50.}  * indicates our reproduced results.}
  \label{tab:resnet}
  \centering
  \begin{tabular}{lllllll}
    \toprule
    \multirow{2}{*}{Model}& \multirow{2}{*}{Params} & \multirow{2}{*}{\#Res} & \multicolumn{2}{c}{Baseline} & \multicolumn{2}{c}{Ours} \\
    \cmidrule(r){4-7}
    &  &   & Top-1  & Top-5 & Top-1  & Top-5\\
    \midrule
    ResNet-50   & 25.6M   & $224^2$ & 78.5*  & 94.4* & 79.0 & 94.7\\
    \bottomrule
  \end{tabular}
\end{table}
We try to extend our proposed training strategy to ResNet-50~\cite{he2016deep}, a commonly used CNN architecture in computer vision. As illustrated in Table~\ref{tab:resnet}, the experimental results show the generality of the proposed training paradigm, which demonstrates that there is still a similar situation in the training process of CNN-based network architecture, \ie, deep neural networks following an easy sample to hard sample learning process, and reasonable manipulation of training samples at different evolutionary periods of models can improve the performance.

\begin{figure}[t]
  \centering
     \begin{subfigure}{0.30\textwidth}
    \includegraphics[width=\linewidth]{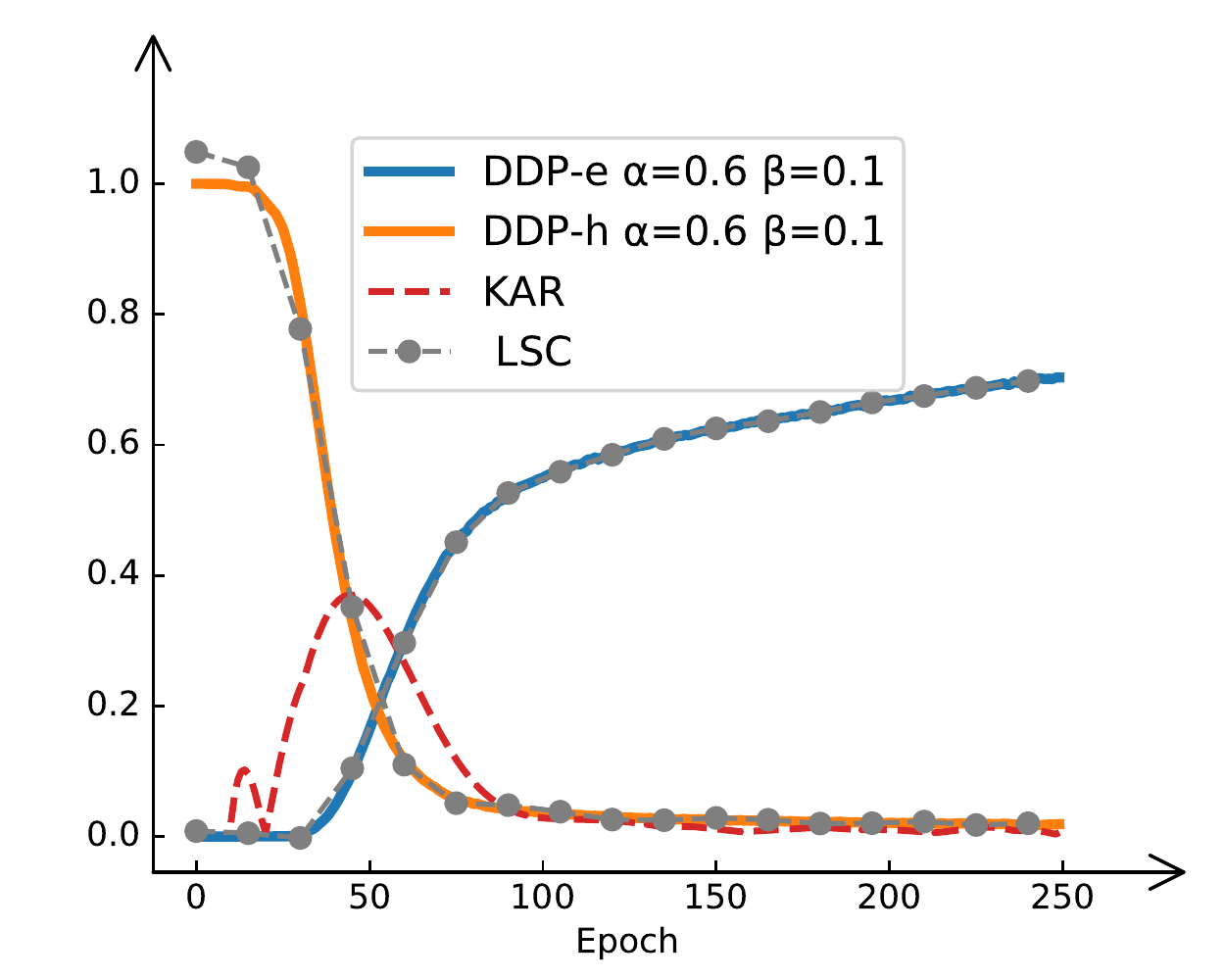}
    \caption{Swin-T: $\alpha$=0.6, $\beta$=0.1.}
    \label{subfig:swin_small}
    \end{subfigure}
    \hfill
    \begin{subfigure}{0.30\textwidth}
    \includegraphics[width=\linewidth]{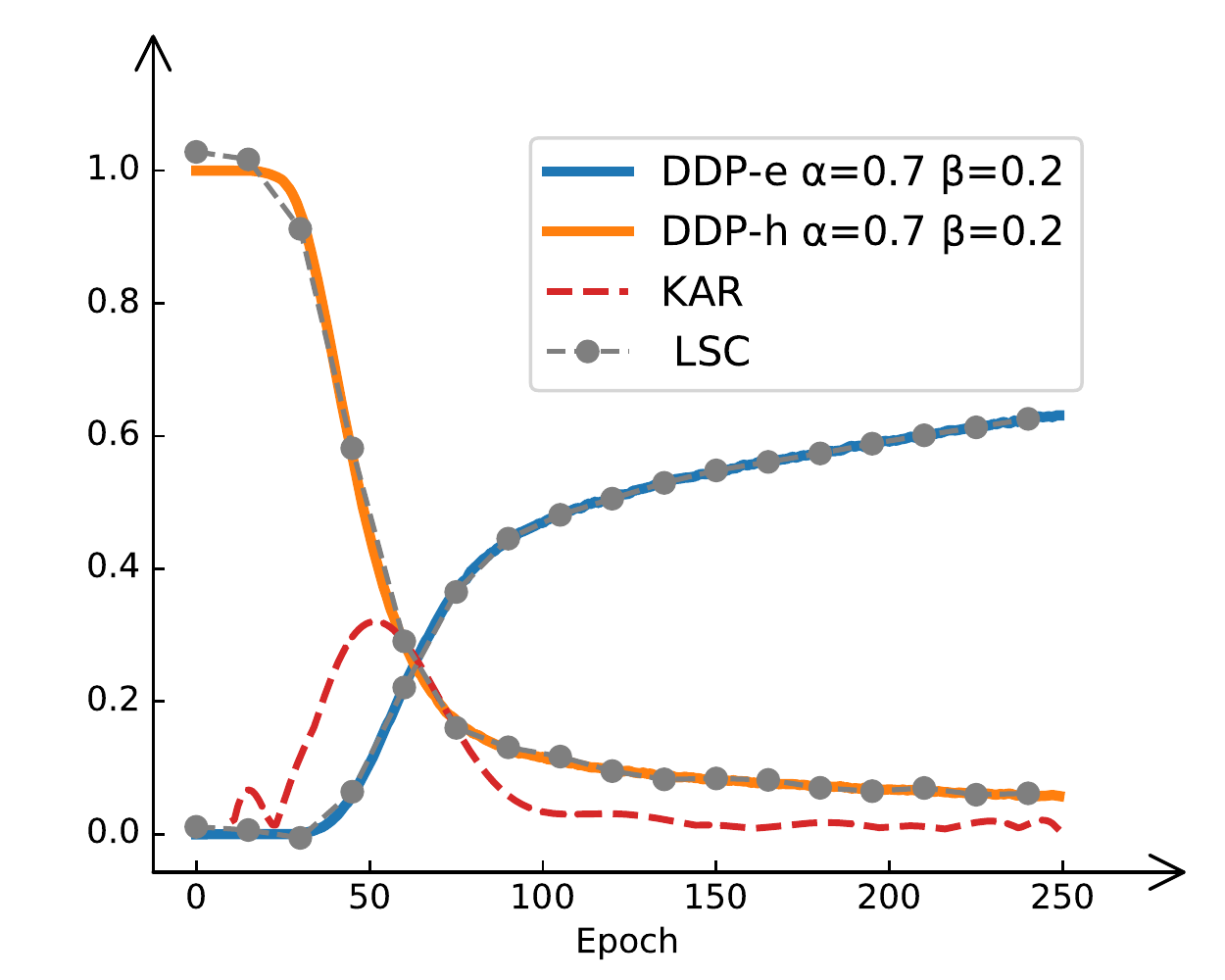}
    \caption{Swin-T: $\alpha$=0.7, $\beta$=0.2.}
    \label{subfig:deit_tiny}
    \end{subfigure}
    \hfill
    \begin{subfigure}{0.30\textwidth}
    \includegraphics[width=\linewidth]{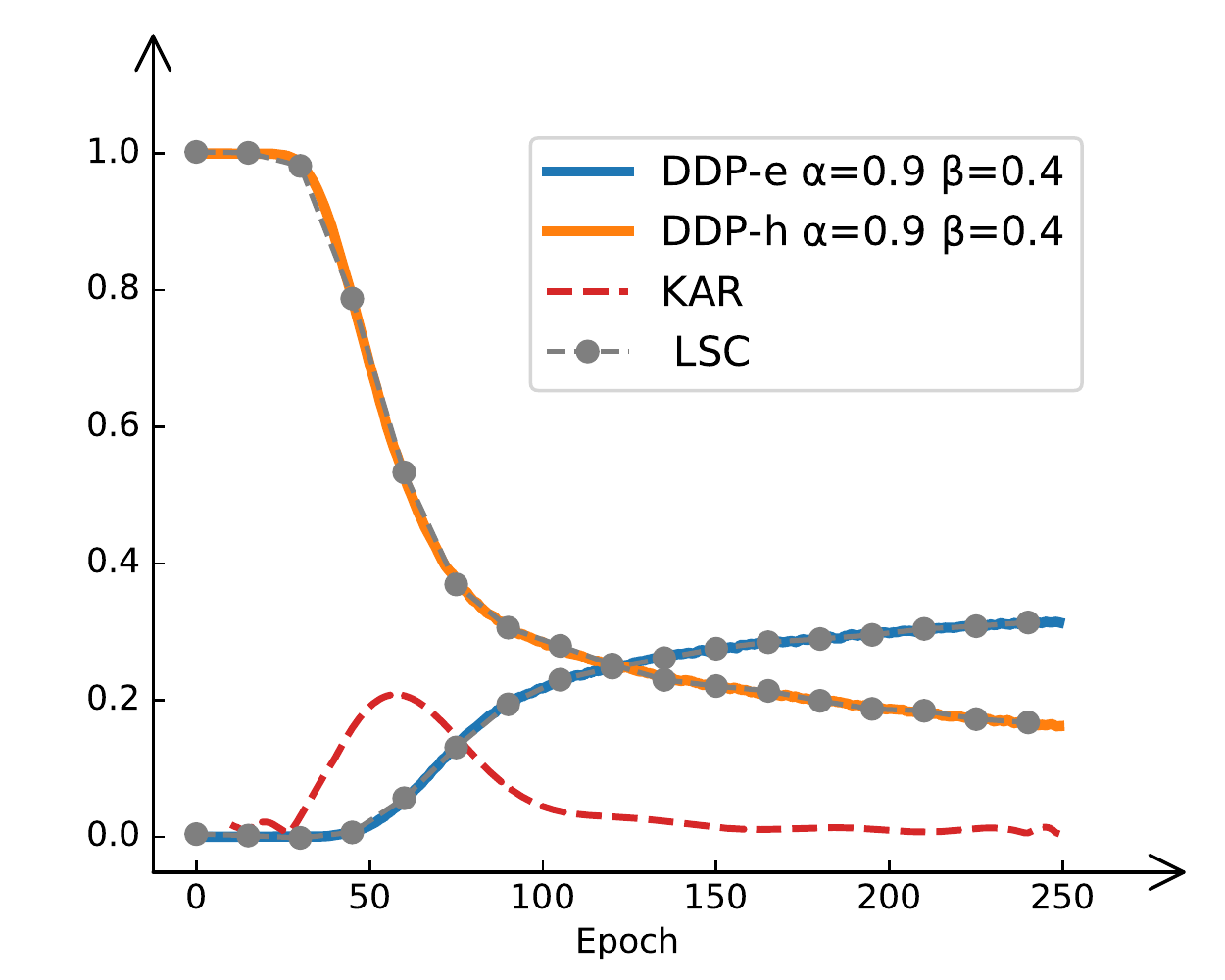}
    \caption{Swin-T: $\alpha$=0.9, $\beta$=0.4.}
    \label{subfig:deit_base}
    \end{subfigure}
    \hfill
    \begin{subfigure}{0.30\textwidth}
    \includegraphics[width=\linewidth]{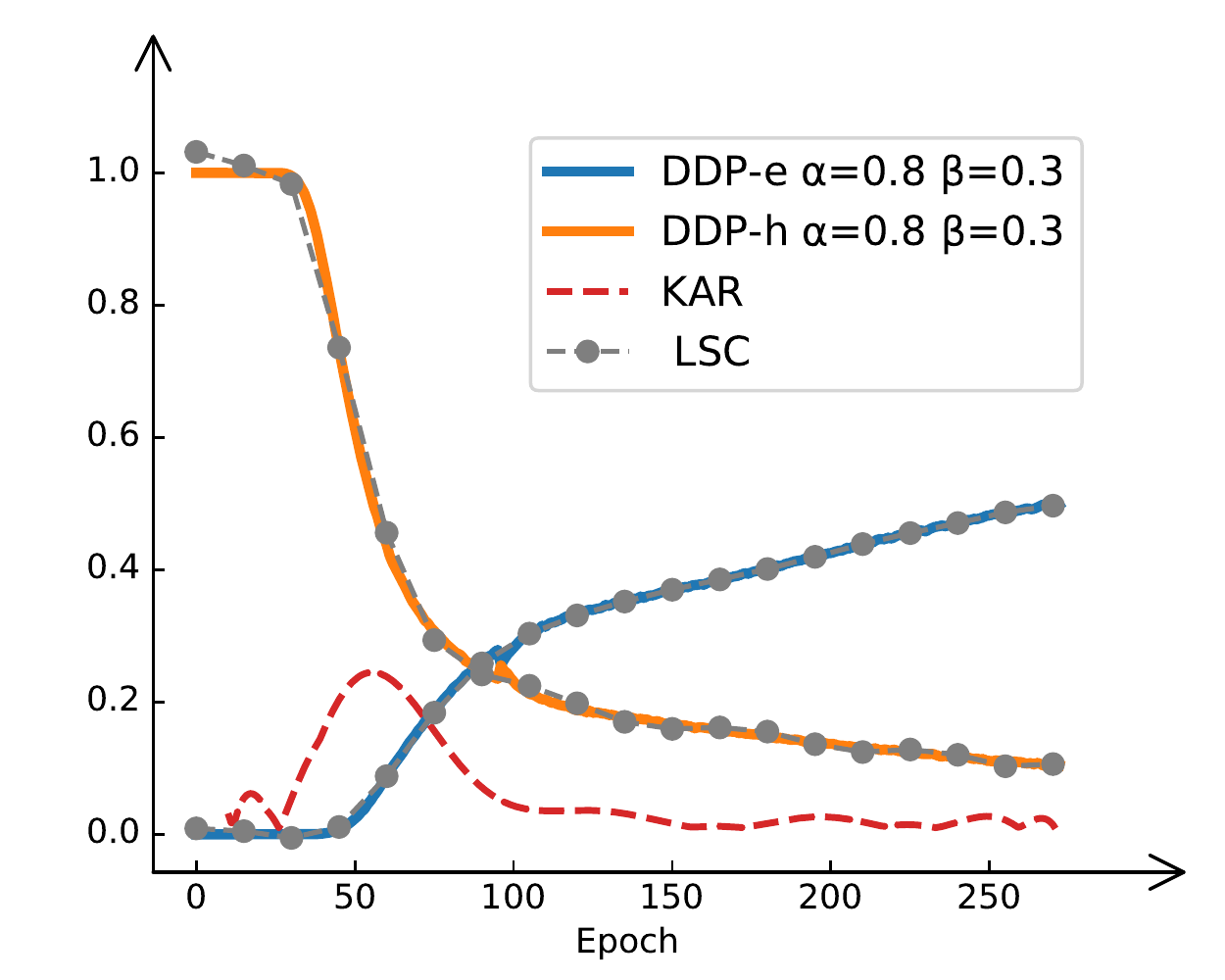}
    \caption{Swin-T: $\alpha$=0.8, $\beta$=0.3.}
    \label{subfig:deit_tiny}
    \end{subfigure}
    \hfill
    \begin{subfigure}{0.30\textwidth}
    \includegraphics[width=\linewidth]{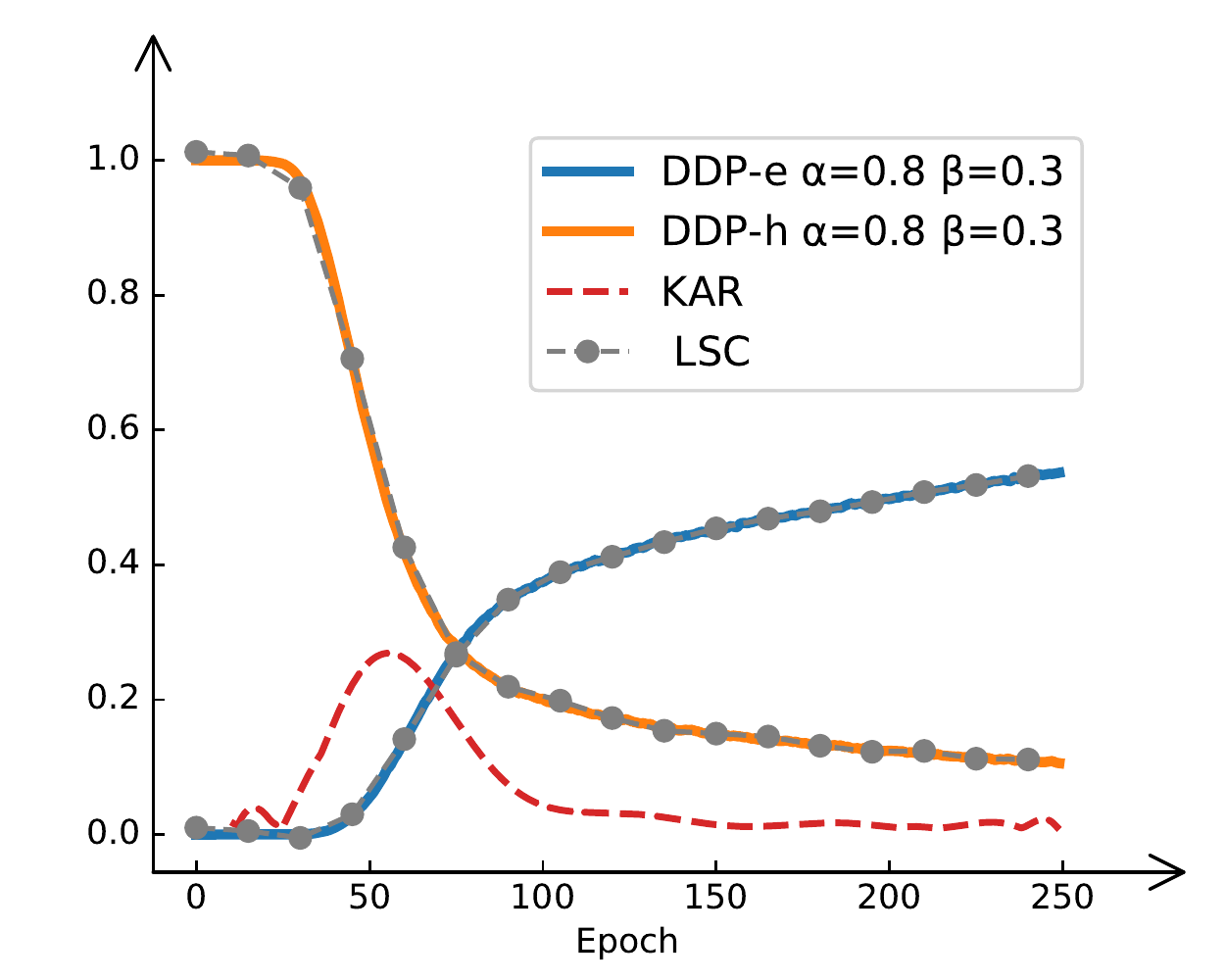}
    \caption{Swin-S: $\alpha$=0.8, $\beta$=0.3.}
    \label{subfig:deit_small}
    \end{subfigure}
    \hfill
    \begin{subfigure}{0.30\textwidth}
    \includegraphics[width=\linewidth]{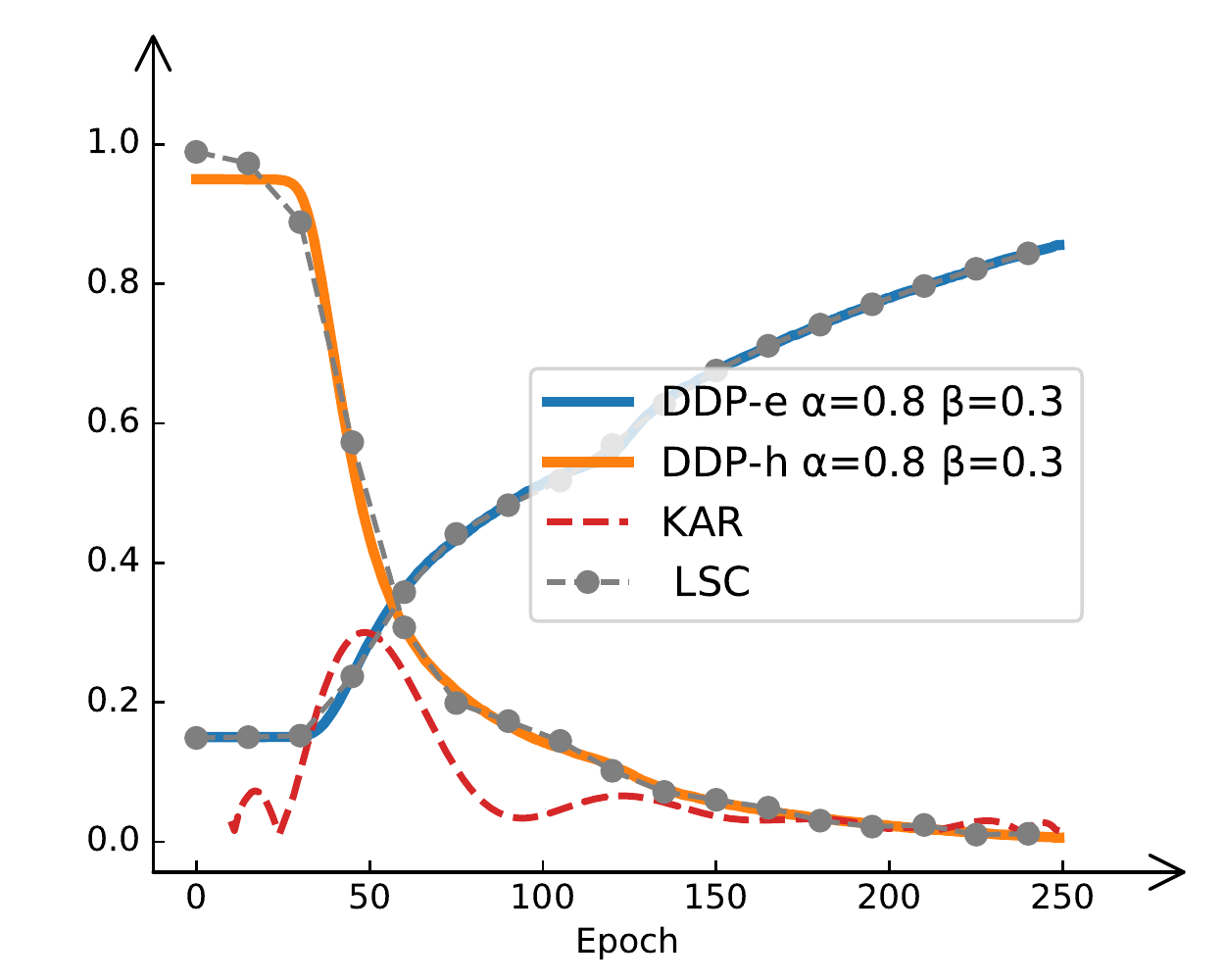}
    \caption{Swin-B: $\alpha$=0.8, $\beta$=0.3.}
    \label{subfig:swin_base}
    \end{subfigure}
  \caption{\textbf{The evolutionary period of Swin Transformer.} Wherein the subfigures (a), (b) and (c) show the DDP and KAR curves under different thresholds $\alpha$ and $\beta$ on Swin-T. The subfigures (d), (e) and (f) show the DDP and KAR curves of models with different capacities under the same thresholds $\alpha$ and $\beta$.}
  \label{fig:swin_period}
\end{figure}

\subsection{Impacts of PatchErasing on Training Data} \label{impt_PE}
\begin{figure}[!t]
    \centering
    \includegraphics[width=0.8\linewidth]{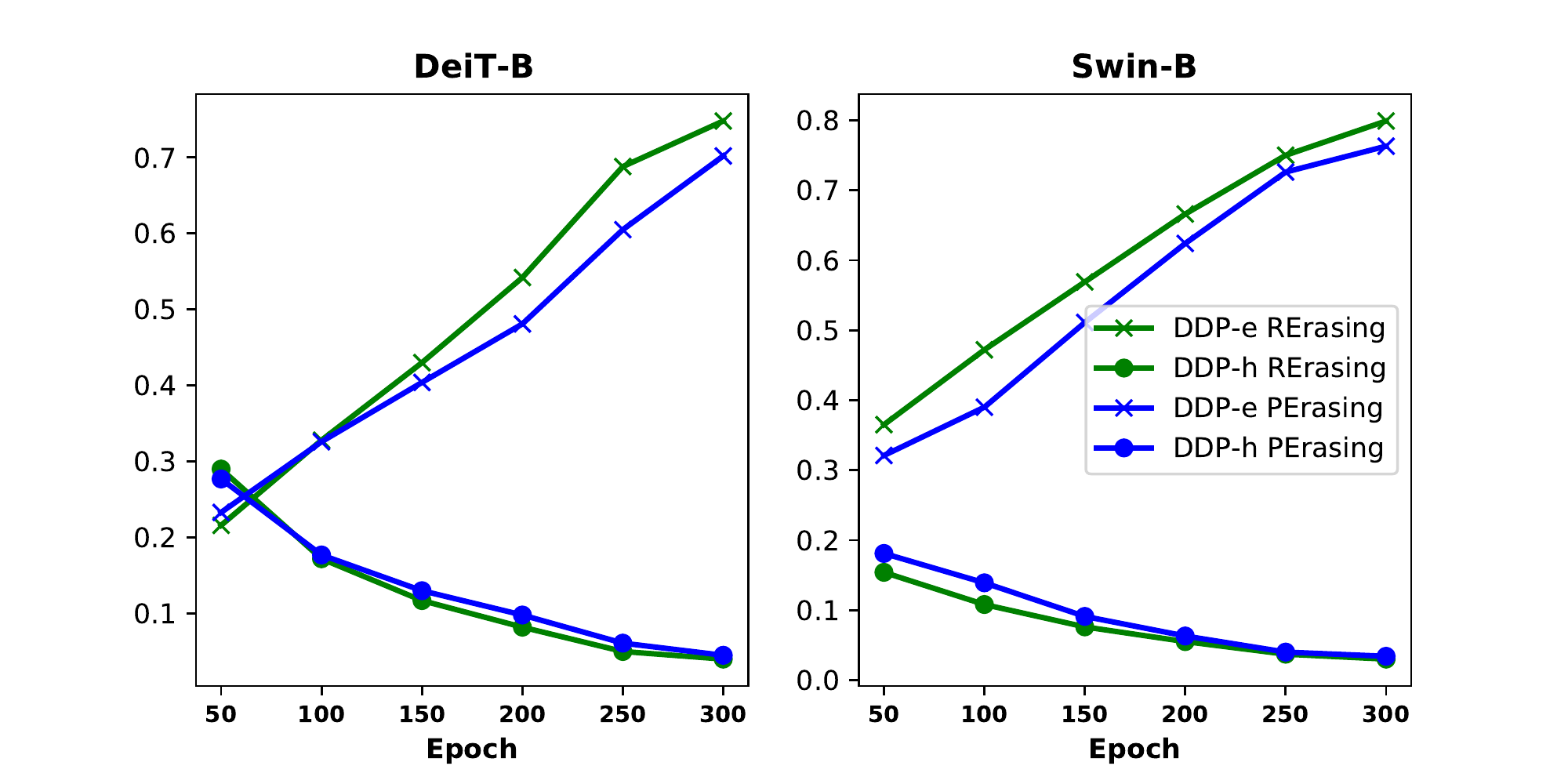}
    \caption{\textbf{The $\mathrm{DDP}_e$ and $\mathrm{DDP}_h$ values under two regularization methods.} The measurements are evaluated from the training set of the ImageNet1K dataset, where all examples here only use the RandomErasing / PatchErasing for data preprocessing. Furthermore, we set the mask rate to $\mu$=30\% for PatchErasing, and take $\alpha$=0.8, $\beta$=0.3 in this experiment.
    }
    \label{fig:e2h_analyze}
\end{figure}
As aforementioned in the main manuscript, PatchErasing is introduced to enlarge the knowledge capacity of training data. To dive into this view more intuitively, we further analyze its effects on the $\mathrm{DDP}_{t,e}$ and $\mathrm{DDP}_{t,h}$ of DeiT-B and Swin-B. In particular, we measure these metrics at different epochs on the ImageNet1K training set. 
As shown in Figure \ref{fig:e2h_analyze}, compared with RandomErasing strategy, PatchErasing can further reduce the number of easy samples ($\mathrm{DDP}_{t,e}$) while increasing the number of hard samples ($\mathrm{DDP}_{t,h}$) during training, which means more ``effective'' examples are generated for networks training. In other words, the PatchErasing can encourage the network to absorb more effective gradients by converting easy samples to hard ones, as we find that easy samples increase significantly late in the $T_3$ period, and they provide very little positive effect in network training.

\subsection{Evolution Curves of Swin Transformer} \label{evolution_curves}
Considering that DeiT-S and Swin-T have a similar number of parameters (22M $vs.$ 28M), we thus show the DDP and KAR curves under different thresholds $\alpha$ and $\beta$ on Swin-T. As shown in Figure\ref{fig:swin_period}(a,b,c), the learning process of Swin Transformer still follows the learning process of three evolution periods $T_1$, $T_2$ and $T_3$, according to the learning speed and ability to assimilate new knowledge. At the same time, we interestingly find that Swin Transformer with different capacities shows nearly the same knowledge assimilation rate (KAR) under the same thresholds $\alpha$ and $\beta$ in the $T_2$ period as shown in Figure\ref{fig:swin_period}(d,e,f), which may be attributed to the stronger local and global modeling ability of Swin Transformer. In other words, the Swin Transformer, which takes into account both local and global modeling, has a stronger learning ability than the vanilla global-based ViT architectures.

\begin{figure}[!ht]
  \centering
    \includegraphics[width=1.0\linewidth]{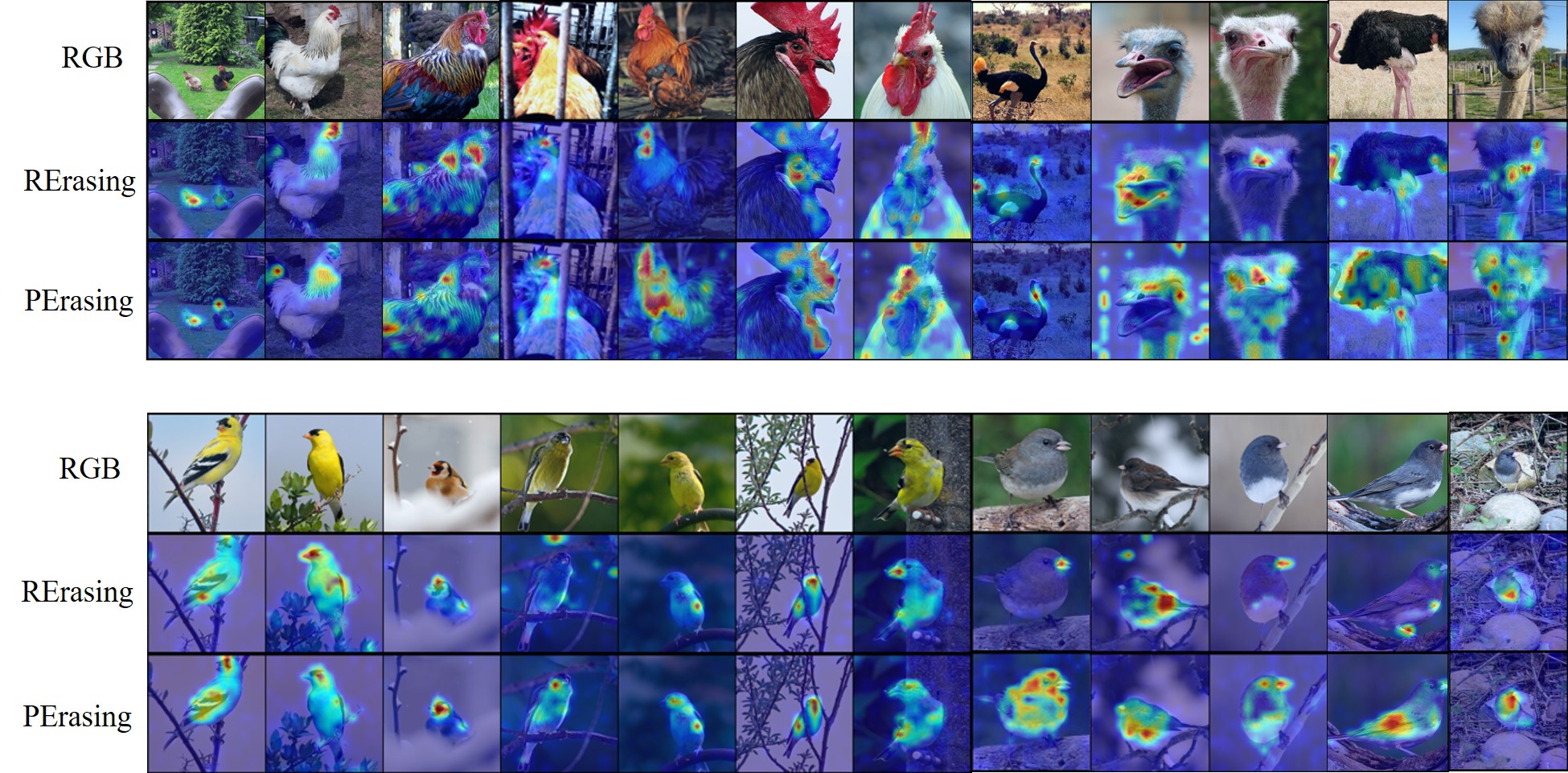}

  \caption{\textbf{Class activation map on ImageNet validation images.}}
  \label{fig:more_visualize}
\end{figure}
\subsection{Visualization of CAM}\label{visualize}
We visualize more class activation maps in Figure \ref{fig:more_visualize}, where the first row represents the original images, the second row represents applying a random erasure strategy during training, and the third row represents applying the proposed patch-level regularization strategy during training.
As can be seen, the patch-level regularization strategy force the network to activate more regions of the object, instead of focusing only on those most sensitive ones. This is beneficial because the network can still be confident enough to recognize instances in the image when the main area of the object is occluded. In addition, we find that the model trained with the PatchErasing strategy is more friendly to these smaller-scale and larger-scale objects.

\end{document}